\documentclass[letterpaper,journal]{IEEEtran}

\usepackage{amsmath,amssymb}
\usepackage{graphicx}
\usepackage{array}
\usepackage{algorithmic}
\usepackage{algorithm}
\usepackage[caption=false,font=footnotesize]{subfig}
\usepackage[nocompress]{cite}
\usepackage{url}

\usepackage[hidelinks]{hyperref}

\ifCLASSINFOpdf
\else
\fi

\begin{document}
\title{TERRA: A Hierarchical Parallel Training and Memory Orchestration Framework for High-Resolution AI-based Earth Modeling}

\author{Ruohan~Wu,
        Ziqi~Zhu,
        Yang~Zhao,
        Jiarui~Tang,
        Yingzhe~Cui,
        Junshi~Chen,
        Zhao~Jing,
        Jun~Shi,
        and~Hong~An\thanks{This work was supported in part by Laoshan National Laboratory
under Grant LSKJ202300305 and in part by the Anhui Provincial Natural
Science Foundation under Grant 2508085QF210.
(Corresponding authors: Jun Shi and Hong An.)}\thanks{Ruohan Wu and Jun Shi are with the School of Computer Science
and Technology, University of Science and Technology of China,
Hefei 230026, China
(e-mail: ruohanwu@mail.ustc.edu.cn;
shijun18@ustc.edu.cn).}\thanks{Ziqi Zhu is with the School of Artificial Intelligence and
Data Science, University of Science and Technology of China,
Hefei 230026, China
(e-mail: ta1ly@mail.ustc.edu.cn).}\thanks{Yang Zhao, Jiarui Tang, Junshi Chen, and Hong An are with the School of Computer Science and Technology, University of Science and Technology of China, Hefei 230026, China, and also with the Department of Ocean Big Data and Artificial Intelligence, Laoshan Laboratory, Qingdao, China
(e-mail: yang\_zhao@mail.ustc.edu.cn;
tjr@mail.ustc.edu.cn;
cjuns@ustc.edu.cn;
han@ustc.edu.cn).}\thanks{Yingzhe Cui is with the Department of Ocean Big Data and Artificial Intelligence, Laoshan Laboratory, Qingdao, China
(e-mail: cuiyingzhe@lsnl.cn).}\thanks{Zhao Jing is with the Frontiers Science Center for Deep Ocean Multispheres and Earth System and the Key Laboratory of Physical Oceanography/Academy of Future Ocean, Ocean University of China, Qingdao, China, and also with the Department of Ocean Big Data and Artificial Intelligence, Laoshan Laboratory, Qingdao, China
(e-mail: jingzhao@ouc.edu.cn).}}

\maketitle

\begin{abstract}
Training high-resolution AI-based Earth forecasting models is memory-intensive. Window-based Swin Transformers reduce the quadratic cost of global attention, but existing distributed systems such as AERIS primarily target pixel-level models and do not jointly support convolutional sampling modules and shifted-window execution. Long-lead rollout finetuning further increases activation memory. To address these challenges, we present TERRA, a hierarchical parallel training framework for high-resolution Earth forecasting. TERRA introduces Sampling-Aware Window, Sequence, and Tensor Parallelism (SAWSTP), which preserves spatially contiguous layouts for sampling modules and routes tokens into topology-aware ragged window layouts for Transformer execution. For long-lead finetuning, Memory Orchestration (MO) provides rollout-aware checkpoint planning and combines input buffering with budget-constrained activation offloading. Experiments on the $1/12^\circ$ GLORYS-based Wenhai workload show that TERRA supports models with up to 11.4B parameters on 96 H200 GPUs and sustains up to $39.76$ PFLOPS, achieving $65.0\%$ strong-scaling and $94.1\%$ weak-scaling efficiency. Compared with checkpoint-only policies, MO further reduces peak allocated GPU memory by $32.2\%$--$51.8\%$ with at most $20.0\%$ step-time overhead, which makes finetuning with smaller patch sizes and longer rollouts feasible for improved forecasting accuracy.
\end{abstract}


\IEEEpeerreviewmaketitle

\section{Introduction}
\IEEEPARstart{A}{I}-based Earth forecasting has emerged as a promising approach for advancing high-resolution weather and ocean prediction~\cite{23fourcastnet,23pangu,23fuxi,23graphcast,25Gencast,25wenhai,25-Aurora}. Recent models increasingly adopt finer spatial resolutions to represent complex Earth-system dynamics. However, training these models at scale remains challenging because high-resolution, multivariate training data increases the number of spatial tokens and creates large activation tensors.

Recent systems have demonstrated that large-scale parallel training can support Earth forecasting models with greater parameter capacity. ORBIT~\cite{24orbit} combines Fully Sharded Data Parallelism (FSDP)~\cite{23fsdp} with tensor parallelism (TP)~\cite{19megatron} to train Vision Transformer (ViT)-based Earth models~\cite{20vit} with up to 113B parameters on $1.4^\circ$ data. At higher spatial resolutions, however, activation memory rather than parameter-related memory can become the dominant training bottleneck. Window-based Swin Transformers~\cite{21swin} are therefore particularly attractive because they avoid the quadratic cost of global attention. AERIS~\cite{25aeris} scales pixel-level Swin Transformers with up to 80B parameters on $0.25^\circ$ data through Sequence-Window Parallelism (SWiPe), which combines window parallelism (WP), sequence parallelism (SP)~\cite{23ulysses}, and pipeline parallelism (PP)~\cite{19gpipe}.

Despite this progress, the SWiPe method for pixel-level Swin Transformers still faces several challenges. First, SWiPe does not support convolutional sampling modules, which reduce the token resolution and activation memory of Transformer blocks and can also improve forecasting accuracy. Second, SWiPe relies on PP, whose bubbles cannot be effectively amortized under small global batch sizes. Third, AERIS considers only fixed window assignments. When the window size is small and the number of windows per rank is sufficient, complete-window WP provides sufficient parallelism, making SP communication unnecessary. When the window size is large, fixed window assignment requires substantial padding to maintain uniform window ownership. Its round-robin ownership also increases the communication required by shifted-window execution. Moreover, using smaller patch sizes~\cite{24-nips-ps} and finetuning with longer lead times can improve long-lead forecasting accuracy. However, these methods also significantly increase activation memory, especially for the $1/12^\circ$ GLORYS-based Wenhai workload~\cite{25wenhai, glorys}.

To address these challenges, we present TERRA, a hierarchical parallel training and memory orchestration framework for high-resolution Earth forecasting. TERRA introduces Sampling-Aware Window, Sequence, and Tensor Parallelism (SAWSTP), which defines an I/O topology for sampling modules and a Transformer topology for Swin Transformer blocks. The I/O topology applies Domain Parallelism (DMP) to partition high-resolution inputs into spatially contiguous regions and parallelizes convolutional sampling modules through halo exchange~\cite{2019-conv-halo}. The Transformer topology integrates WP, SP, and TP with FSDP, avoiding PP and its pipeline bubbles at small global batch sizes. Differentiable token routes connect the two topologies by converting spatial DMP partitions into the window-oriented layouts required by Transformer blocks and support distributed shifted-window execution across different WP, SP, and TP configurations.

Building on these token routes, TERRA introduces cost-aware window assignment to adapt the Transformer topology to different window sizes and parallel configurations. It uses ragged window ownership and selects among WP/SP topologies to reduce padding and shifted-window communication while balancing workloads. 

To reduce the activation-memory pressure of long-lead finetuning, TERRA further introduces Memory Orchestration (MO). Building on activation checkpointing~\cite{16-chen-ckpt,16bptt,20dtr,23rockmate}, MO separately profiles the retained activation memory and backward recomputation peaks of sampling and Transformer checkpoint units. It uses these measurements to construct a lead-dependent checkpoint plan that balances rollout-wide activation accumulation against local recomputation peaks. Input buffering bounds the GPU lifetime of rollout inputs and labels. If the predicted peak still exceeds the target memory budget, MO selectively offloads activation boundaries to host memory~\cite{16vdnn,19vdnn++,21-nips-ckpt,24delta} until the predicted peak satisfies the budget.

We implement TERRA as an open-source training framework available at
\url{https://github.com/ruohan12345/TERRA} and evaluate it through
pretraining and rollout finetuning of hierarchical GLORYS-based Wenhai
models. The contributions of this work are summarized as follows:
\begin{itemize}
  \item We design SAWSTP, a parallelism strategy that applies halo-exchange-based DMP to convolutional sampling modules and combines WP, SP, TP, and FSDP for Transformer blocks. It scales Wenhai model training to 96 H200 GPUs, achieving up to $65.0\%$ strong-scaling and $94.1\%$ weak-scaling efficiency, with sustained performance of up to $39.76$~PFLOPS.

  \item We introduce cost-aware ragged window assignment that adapts the WP/SP topology to the window size and parallel configuration. It reduces padding and shifted-window communication while balancing workloads, providing a $1.03\times$--$1.18\times$ training-step speedup over the fixed AERIS-style baseline.

  \item We develop MO for long-lead finetuning by combining profile-guided, lead-dependent checkpoint planning with input buffering and budget-constrained activation offloading. MO reduces peak allocated GPU memory by $32.2\%$--$51.8\%$ relative to checkpoint-only policies, with at most $20.0\%$ step-time overhead.

  \item We pretrain, finetune, and evaluate hierarchical Wenhai models on the GLORYS-based workload. The results demonstrate that hierarchical sampling, smaller patch sizes, and longer rollout finetuning enabled by TERRA improve long-lead forecasting accuracy.
\end{itemize}

\section{Background and Motivation}\label{sec:background_and_moti}

\subsection{AI-based Earth Forecasting Models}
\begin{figure*}[!t]
\centering
\subfloat[Hierarchical Swin-based model.]{
  \includegraphics[width=0.34\textwidth]{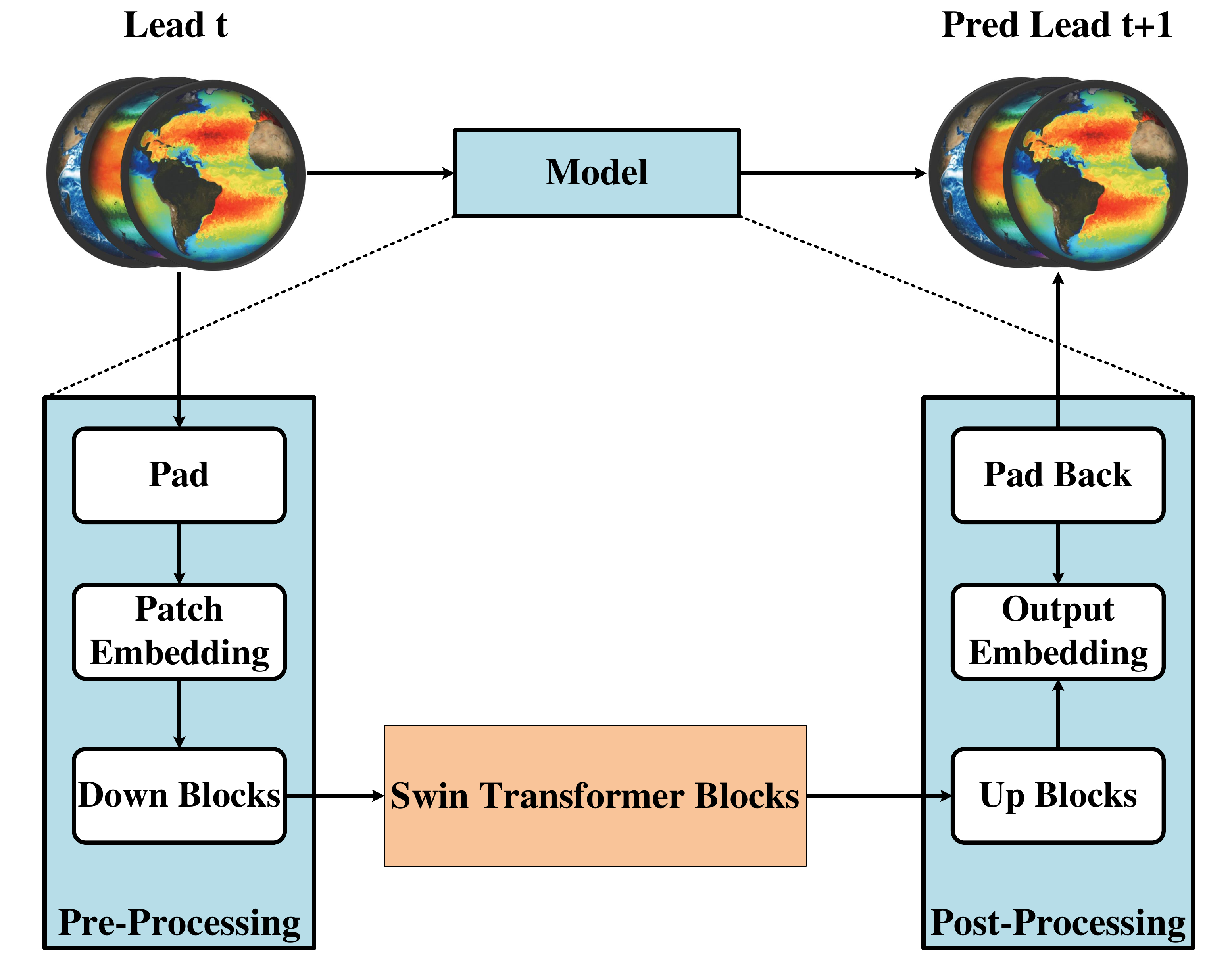}
  \label{fig:pt_ft_a}
}
\hfil \subfloat[One-step pretraining and multi-step rollout finetuning.]{
  \includegraphics[width=0.48\textwidth]{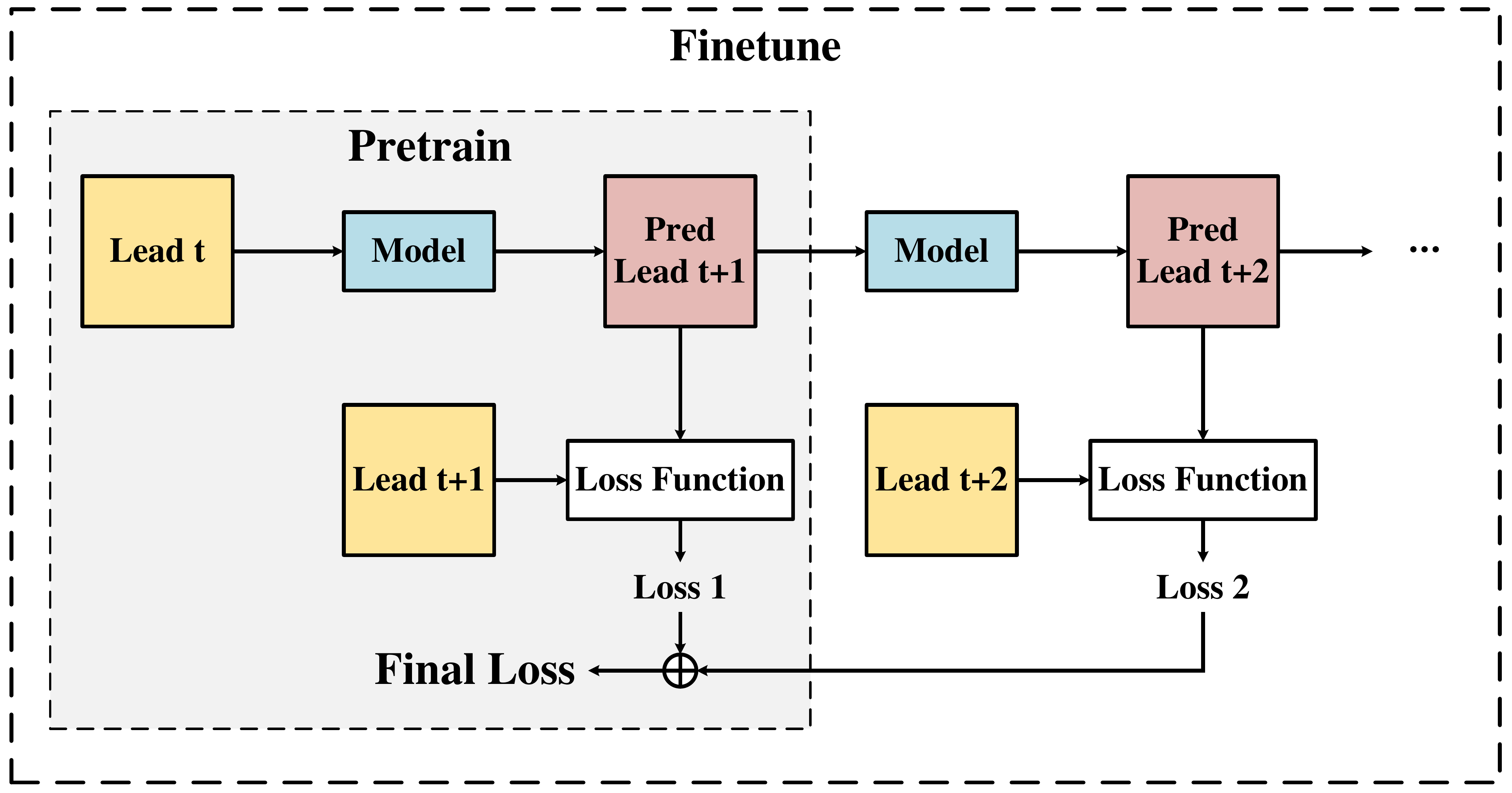}
  \label{fig:pt_ft_b}
}
\caption{High-resolution Earth forecasting model and training stages.}
\label{fig:pt_ft}
\end{figure*}

AI-based models have become an important approach for global weather and ocean forecasting~\cite{23fourcastnet,23pangu,23fuxi,23graphcast,25Gencast,25wenhai,25-Aurora}. However, the increasing spatial resolution substantially raises the computational and memory requirements of model training. For example, Pangu-Weather is trained on $1/4^\circ$ ERA5 data with $721\times1440$ grid points~\cite{23pangu,era5}, whereas Wenhai uses $1/12^\circ$ GLORYS data with $2041\times4320$ grid points~\cite{25wenhai,glorys}. TERRA targets the Wenhai workload, whose high-resolution multivariate ocean inputs impose substantial memory pressure during training.

In Wenhai, each input is a dense multivariate ocean field on a regular latitude--longitude grid, represented as $X_t\in\mathbb{R}^{B\times H\times W\times C}$, where $B$ is the micro-batch size, $H$ and $W$ are the spatial dimensions, and $C=93$ is the number of ocean variables. As shown in Fig.~\ref{fig:pt_ft}\subref{fig:pt_ft_a}, Wenhai first maps the input field to patch tokens, which are then processed by a hierarchical encoder--decoder architecture that combines convolutional down- and up-sampling modules with window-based Swin Transformer blocks~\cite{21swin}. This design reduces the number of tokens processed by the Transformer blocks while preserving multiscale spatial representations. Table~\ref{tab:notation} summarizes the main notation of the Wenhai model.

\begin{table}[!t]
\centering
\caption{Notation for the Wenhai model.}
\label{tab:notation}
\footnotesize
\renewcommand{\arraystretch}{1.08}
\setlength{\tabcolsep}{4pt}
\begin{tabular}{c|p{0.68\columnwidth}}
\hline
\textbf{Notation} & \textbf{Description} \\ \hline
$H,W$ & Spatial resolution of the input field \\
$\hat{H},\hat{W}$ & Padded spatial resolution of the input field \\
$C$ & Number of input physical variables \\
$p$ & Patch size \\
$\alpha$ & Stride of each down- or up-sampling module \\
$d$ & Embedding dimension \\
$N_{\mathrm{h}}$ & Number of attention heads \\
$w$ & Attention-window size \\
$L$ & Number of Swin Transformer blocks \\
\hline
\end{tabular}
\end{table}

As shown in Fig.~\ref{fig:pt_ft}\subref{fig:pt_ft_b}, AI-based Earth forecasting is autoregressive. Given an initial state $X_t$ at time $t$, the predicted state at lead time $k\Delta t$ is fed back as the input to the next forecast step, producing the state at lead time $(k+1)\Delta t$. To improve long-lead forecast accuracy, models are typically pretrained for single-step forecasting and then finetuned over multiple autoregressive steps~\cite{23fourcastnet,23fuxi,23graphcast,25wenhai}. During pretraining, the model learns a single-step forecasting operator $f_\theta$ that maps $X_t$ to $X_{t+\Delta t}$. During finetuning, the same model parameters are reused over a $K$-step rollout:
\begin{equation}
\hat{X}_{t+k\Delta t} =
f_\theta\!\left(\hat{X}_{t+(k-1)\Delta t}\right),
\qquad 1\leq k\leq K,
\end{equation}
where $\hat{X}_t=X_t$ and $K$ is the rollout length. The finetuning objective sums the prediction losses over the $K$ forecast steps, which directly optimizes long-lead forecast accuracy. However, the finetuning also retains intermediate activations across the rollout until backpropagation, which substantially increase the memory footprint.

\subsection{Memory-Efficient Parallel Training}
Training high-resolution forecasting models presents both substantial I/O overhead and memory pressure. A single training sample can contain a large multivariate spatial field, making the micro-batch size per GPU typically small (e.g. $B=1$). Recent systems therefore distribute one sample across multiple devices to reduce the per-rank input and activation footprint~\cite{24orbit,25jigsaw, 25orbit2, 25aeris, 26shardtensor}.

The GPU memory footprint of DNN training consists primarily of model parameters, gradients, optimizer states, and activations~\cite{22patrickstar}. Different parallel and memory-management techniques target different components. Data parallelism (DP) replicates the model and distributes samples across ranks, but it cannot reduce the memory required by an individual high-resolution sample. FSDP and ZeRO-style methods shard parameters, gradients, and optimizer states~\cite{20zero, 21zero-offload, 21zero-infinity,23fsdp,26superoffload}. TP~\cite{19megatron, 23-mlsys-megatron} partitions Transformer parameters and intermediate activations, SP~\cite{23ulysses} partitions token sequences, and PP partitions model layers into stages~\cite{19gpipe}. Activation checkpointing trades additional recomputation for a smaller activation footprint~\cite{16-chen-ckpt,16bptt,20dtr,23rockmate}, while activation offloading transfers selected activations outside GPU memory for later reuse~\cite{16vdnn,19vdnn++,21-nips-ckpt,24delta}.

To improve the efficiency of large-scale Earth modeling, a series of memory-efficient parallel training systems has been proposed. For example, Jigsaw focuses on spatially partitioned execution for high-resolution inputs through distributed patch embedding and matrix operations~\cite{25jigsaw}. ORBIT combines FSDP and tensor parallelism to scale the parameter capacity of ViT-based Earth foundation models~\cite{24orbit}, while ORBIT-2 further targets hyper-resolution climate downscaling with spatially partitioned execution of extremely long token sequences~\cite{25orbit2}. AERIS addresses distributed training of pixel-level Swin Transformers through SWiPe, which partitions window activations and distributes the computation of shifted-window attention blocks across GPUs~\cite{25aeris}.

\subsection{Motivation}

The following observations motivate the design of TERRA for high-resolution hierarchical Earth forecasting models.

\begin{figure}[!t]
\centering
\subfloat[Patch-size memory.]{
  \includegraphics[width=0.8\columnwidth]{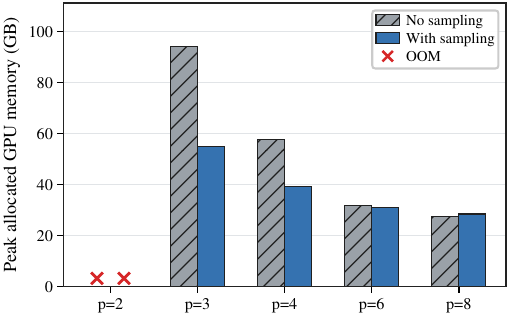}
  \label{fig:motivation_patch_size}
}

\vspace{0.6em}

\subfloat[Rollout memory.]{
  \includegraphics[width=0.8\columnwidth]{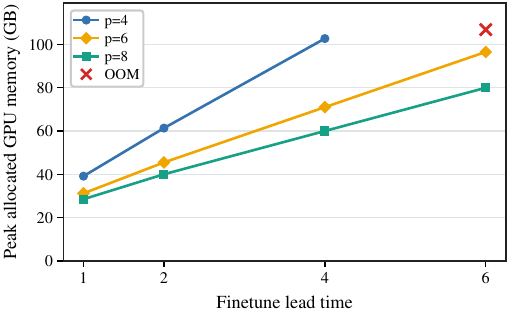}
  \label{fig:motivation_rollout_length}
}
\caption{Memory pressure in training high-resolution Earth forecasting models.}
\label{fig:motivation_mem}
\end{figure}

\subsubsection{Limitations of Existing Swin Parallelism for High-Resolution Earth Modeling and Parameter Scaling}

Prior work has shown that larger models and smaller patch sizes can improve weather-forecasting accuracy~\cite{24-nips-ps}. Existing distributed Swin Transformer systems such as AERIS use SWiPe to train pixel-level Swin Transformers with a patch size of $p=1$ and scale models to 80B parameters~\cite{25aeris}. However, extending this design to higher-resolution training data and more parameters presents additional challenges.

Finer spatial resolutions and smaller patch sizes both increase the number of patch tokens and the associated activation memory. AERIS is evaluated on $0.25^\circ$ data, whereas Wenhai uses $1/12^\circ$ data. At a fixed patch size, this resolution difference produces roughly nine times more tokens and activations. As shown in Fig.~\ref{fig:motivation_mem}\subref{fig:motivation_patch_size}, reducing the patch size substantially increases peak memory, while sampling modules alleviate this pressure by reducing the token resolution processed by the Transformer blocks and can improve forecasting accuracy through multi-scale feature learning. However, combining parallel Transformer execution with parallel sampling modules remains challenging because they favor different tensor layouts. Sampling modules require spatially contiguous partitions for convolution, whereas Swin Transformer blocks require window-major token layouts for window attention.

\begin{table*}[!t]
\centering
\caption{Comparison of model settings and parallelism support in
ORBIT~\cite{24orbit}, AERIS~\cite{25aeris}, and TERRA.}
\label{tab:parallel_dimensions}
\footnotesize
\renewcommand{\arraystretch}{1.15}
\begin{tabular*}{\textwidth}{
@{\extracolsep{\fill}}lccccccccc}
\hline
&
\multicolumn{2}{c}{\textbf{Model setting}} &
\multicolumn{7}{c}{\textbf{Parallelism support}}
\\
\cline{2-3}\cline{4-10}
\textbf{Framework} &
\textbf{Model architecture} &
\textbf{Resolution} &
\textbf{Parallel sampling} &
\textbf{WP} &
\textbf{SP} &
\textbf{TP} &
\textbf{PP} &
\textbf{DP} &
\textbf{FSDP}
\\
\hline
ORBIT &
Vision Transformer &
$1.4^\circ$ &
$\times$ &
$\times$ &
$\times$ &
$\checkmark$ &
$\times$ &
$\checkmark$ &
$\checkmark$
\\
AERIS &
Pixel-level Swin Transformer &
$0.25^\circ$ &
$\times$ &
$\checkmark$ &
$\checkmark$ &
$\times$ &
$\checkmark$ &
$\checkmark$ &
$\times$
\\
TERRA &
Hierarchical Swin Transformer &
$1/12^\circ$ &
$\checkmark$ &
$\checkmark$ &
$\checkmark$ &
$\checkmark$ &
$\times$ &
$\checkmark$ &
$\checkmark$
\\
\hline
\end{tabular*}
\end{table*}

For parameter scaling, Table~\ref{tab:parallel_dimensions} summarizes the model settings and supported parallelism of ORBIT, AERIS, and TERRA. AERIS relies on PP to partition Transformer layers across devices. Its WP and SP distribute activation across multiple GPUs but do not reduce parameter-related memory~\cite{25aeris}. PP bubbles are difficult to amortize under small global batch sizes. For example, Wenhai uses a global batch size of only 16. ORBIT instead combines TP with FSDP to scale Earth foundation models~\cite{24orbit}. However, complex distributed window-attention operations with shifted windows make TP integration with SWiPe challenging.

\subsubsection{Efficient Window Assignment Requires Topology-Aware Optimization}
To support distributed shifted-window execution, SWiPe adopts a fixed window-assignment strategy that maps attention windows to ranks through inter-window WP and intra-window SP. However, this fixed mapping does not adapt to the number of complete windows available under a given parallel topology. When many complete windows are available per rank, the fixed WP mapping can incur avoidable WP communication during distributed window shifting, as well as padding overhead from its regular layout. TERRA mitigates both sources of overhead through a row-contiguous window-assignment method.

When only a few complete windows are available per rank, WP may not divide the window grid evenly. Padding the window grid to equalize window counts increases computation and memory consumption, whereas avoiding padding can leave ranks with imbalanced workloads. Although SP can mitigate this imbalance by further partitioning individual windows across ranks, it introduces additional all-to-all communication. Therefore, window assignment should jointly consider padding overhead, window-shift communication, workload balance, and the required topology of WP and SP.

\subsubsection{Long-Lead Finetuning Requires Activation-Aware Memory Management}

Long-lead finetuning fundamentally changes the memory characteristics of high-resolution Earth-model training. During one-step pretraining, activations are bounded within a single forward--backward pass. In contrast, rollout finetuning repeatedly reuses the same model parameters across multiple forecast steps and retains intermediate rollout states and activations until backpropagation, making long-lead finetuning increasingly dominated by activation memory. As shown in Fig.~\ref{fig:motivation_mem}\subref{fig:motivation_rollout_length}, peak memory increases rapidly with rollout lead time and can eventually cause out-of-memory (OOM) failures.

Although activation checkpointing and activation offloading have been extensively studied for DNN training, existing policies do not adequately capture the heterogeneous memory behavior of sampling and Transformer checkpoint units in hierarchical Earth models. High-resolution sampling blocks retain individually large activation tensors, which are difficult to mitigate through activation checkpointing alone. Moreover, long-lead finetuning requires high-resolution input and label tensors from multiple lead times to reside on the GPU. These tensors alone impose a substantial memory footprint. Consequently, activation-memory optimization should employ separate strategies for different activation units while accounting for the rollout process and remaining compatible with distributed execution, rather than relying on a single global checkpointing or offloading strategy.

\section{TERRA Overview}\label{sec:terra_overview}

\begin{figure}[!t]
\centering
\includegraphics[width=\columnwidth]{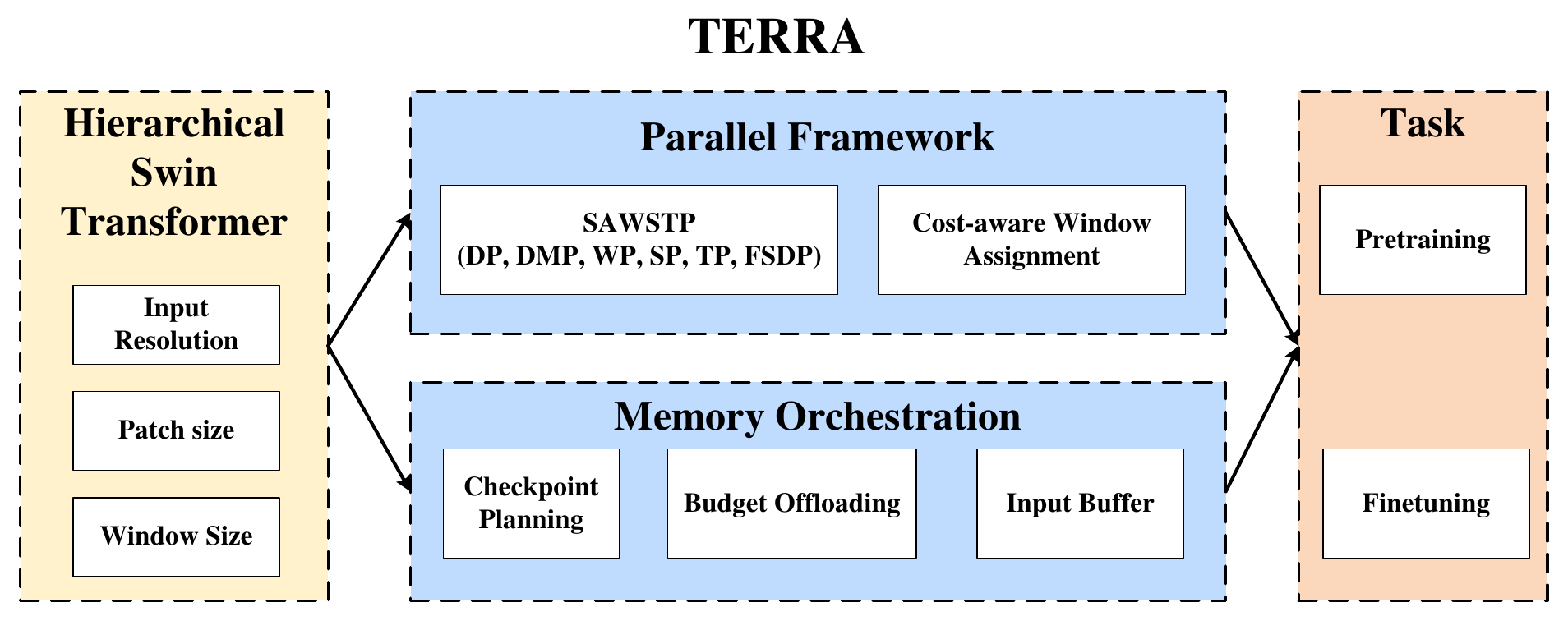}
\caption{Overview of TERRA.}
\label{fig:terra_overview}
\end{figure}

To address the aforementioned challenges, we propose TERRA, a hierarchical parallel training and memory-orchestration framework for high-resolution AI-based Earth forecasting models. As shown in Fig.~\ref{fig:terra_overview}, TERRA targets hierarchical Swin Transformer models and supports configurations with varying input resolutions, patch sizes, window sizes, and model scales. It integrates SAWSTP with MO to support distributed pretraining and long-lead finetuning under these configurations.

SAWSTP partitions each high-resolution training sample into spatially contiguous DMP shards for patch embedding and sampling modules, and then routes the resulting tokens to Swin Transformer blocks executed with WP, SP, TP, and FSDP. Building on SAWSTP, TERRA employs cost-aware window assignment to select a ragged window assignment based on the input and parallel configuration. During long-lead finetuning, MO manages the per-rank memory pressure arising from rollout activations and high-resolution input and label data while preserving the execution semantics of SAWSTP.

\section{Parallel Training Framework}

\begin{figure*}[!t]
\centering
\includegraphics[width=\textwidth]{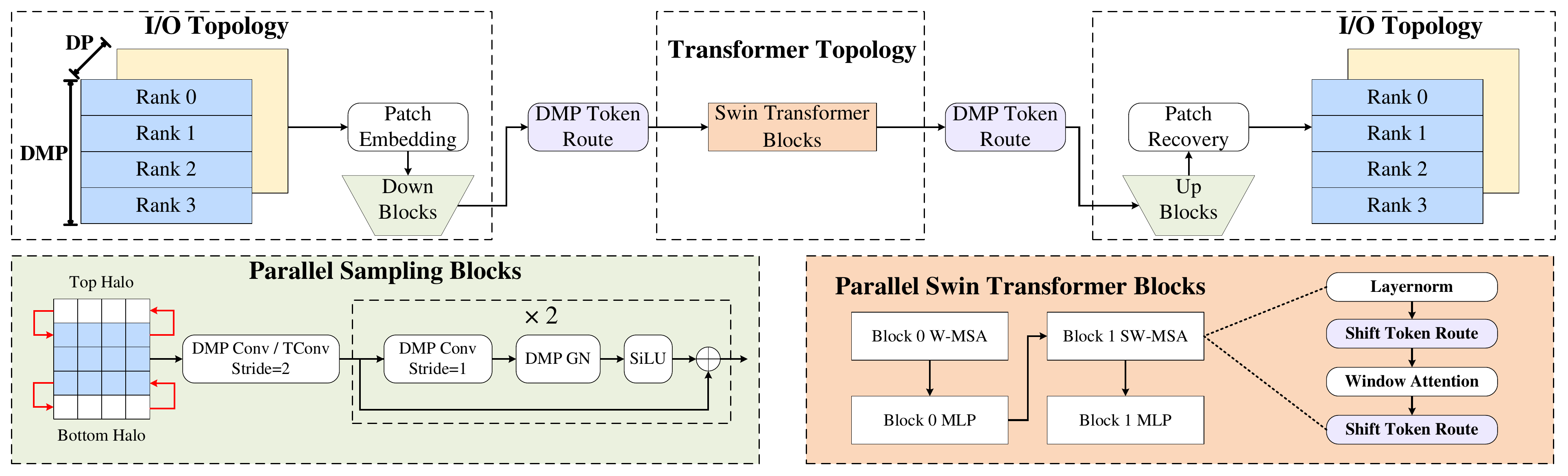}
\caption{SAWSTP Training Framework for the hierarchical Swin Transformer.}
\label{fig_parallel_framework}
\end{figure*}

\subsection{Hierarchical Swin Parallelism}\label{sec:sawstp}

As shown in Fig.~\ref{fig_parallel_framework}, TERRA realizes distributed execution of hierarchical Swin Transformer models through SAWSTP. Hierarchical Swin Transformer models combine convolutional down- and up-sampling modules with Swin Transformer blocks that contain window-based multi-head self-attention (W-MSA) and shifted window-based multi-head self-attention (SW-MSA)~\cite{21swin}. Sampling modules operate on spatially contiguous feature maps to preserve local convolutional dependencies, whereas Transformer blocks require window-major token layouts for distributed attention and shifted-window execution.

To overcome this mismatch, SAWSTP employs two stage-specific logical topologies, namely a sampling-aware I/O topology and a multi-level Transformer topology. The two topologies use the same participating ranks but organize them into different communication groups.

The I/O topology is built on DMP, which partitions each high-resolution training sample and its corresponding labels into spatially contiguous shards. Each rank therefore loads and processes only the input and label tensors that it owns. This topology preserves the DMP layout through patch embedding, sampling, patch recovery, and loss computation.

The Transformer topology remaps the sampled feature maps into window-major layouts for Swin Transformer execution. It factorizes Transformer computation with WP, SP, and TP, which determine complete-window ownership, intra-window sequence partitioning, and operator-level sharding, respectively. FSDP is orthogonal to these parallel dimensions and can shard parameters, gradients, and optimizer states over the combined $\mathrm{DP}\times\mathrm{WP}\times\mathrm{SP}$ process group. The differentiable DMP token route connects the two topologies by converting between DMP-partitioned and window-major layouts. Within the Transformer topology, the shift token route redistributes tokens for shifted-window attention. During backpropagation, the inverse routes ensure correct gradient propagation.

\subsection{Sampling-Aware I/O topology}
The I/O topology first pads the input field $X_t\in\mathbb{R}^{B\times H\times W\times C}$ to a padded shape of $B\times\hat{H}\times\hat{W}\times C$, where $\hat{H}$ and $\hat{W}$ satisfy the divisibility requirements of patch embedding, sampling modules, and subsequent window formation. It then partitions the padded input into $P_g$ contiguous spatial stripes along the height dimension, where $P_g$ is the number of ranks in the DMP group. Each rank owns a local tile and directly loads only its corresponding input and label tensors.

Given a patch size $p$, each rank applies patch embedding to its local tile, producing a feature map of shape $B\times\frac{\hat{H}}{P_g p}\times\frac{\hat{W}}{p}\times d$, where $d$ is the embedding dimension. This feature map serves as the input to the hierarchical sampling modules. Down-sampling and up-sampling modules use strided convolution (Conv) and transposed convolution (TConv), respectively, with stride $\alpha=2$ in this work. Each sampling module also contains residual blocks composed of Conv, Group Normalization (GN), and SiLU operators. Consequently, a down-sampling module reduces the local feature-map shape to $B\times\frac{\hat{H}}{\alpha P_g p}\times\frac{\hat{W}}{\alpha p}\times d$, while the corresponding up-sampling module restores the spatial resolution. The down-sampled features are subsequently routed to the Transformer blocks, substantially reducing Transformer-stage activation memory. This encoder--decoder sampling design also supports multi-scale feature learning, and the evaluation in Fig.~\ref{fig:exp_rmse}\subref{fig:rmse_no_ft} confirms its benefit to forecasting accuracy.

Parallel convolution in sampling modules has been extensively studied in prior training systems~\cite{26shardtensor,2019-conv-halo,19-conv,20-conv,26_ncar_miles_credit}. SAWSTP adopts a halo-exchange-based implementation~\cite{26_ncar_miles_credit} that partitions activations along the height dimension while keeping the Conv, TConv, and GN parameters replicated across DMP ranks. Before each local Conv or TConv operation, neighboring ranks exchange only the boundary rows required by the operator. The communication volume scales with the halo width rather than the full local feature-map size. The halo exchange is differentiable, allowing boundary gradients to be returned to their original ranks during backpropagation.

\subsection{Multi-Level Transformer topology}
\begin{figure*}[!t]
\centering
\includegraphics[width=0.98\textwidth]{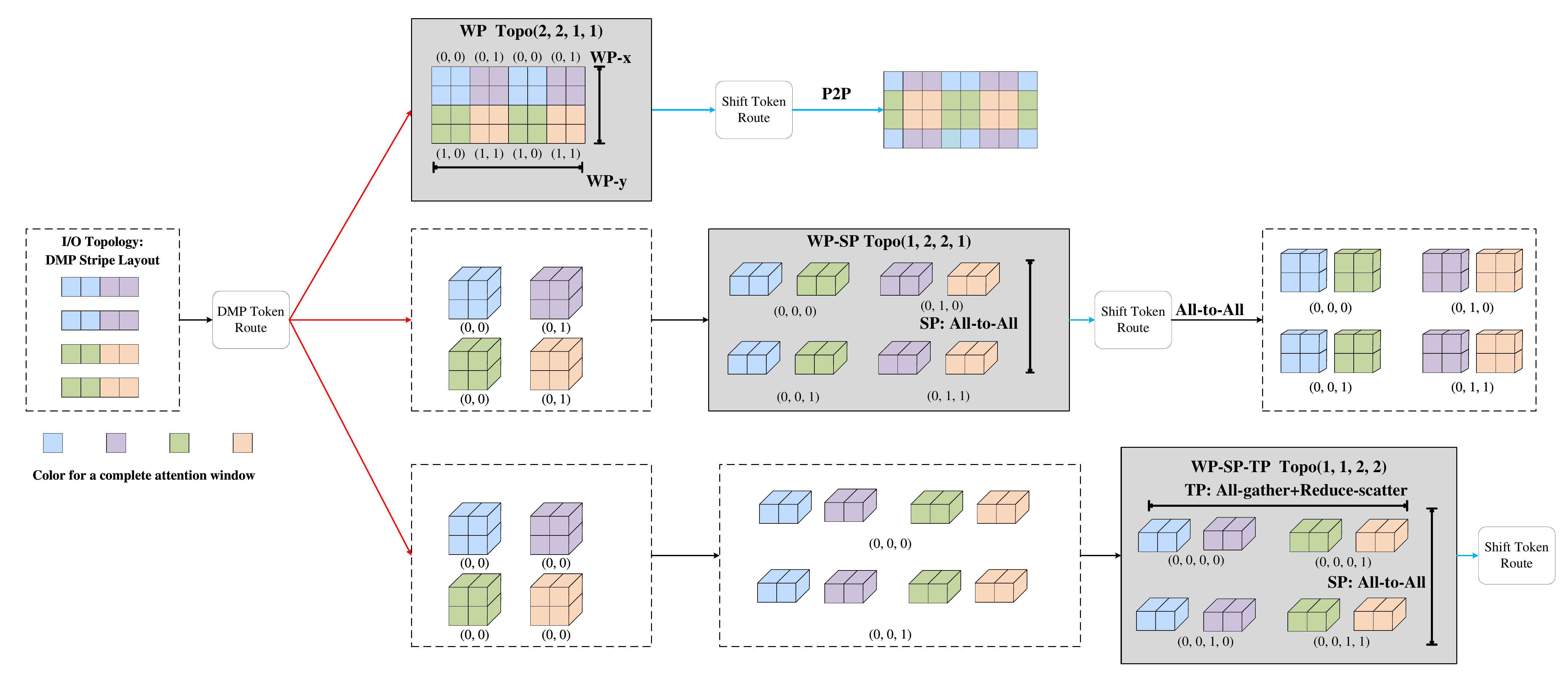}
\caption{Multi-level Transformer topology with DMP and shift token routes in SAWSTP.}
\label{fig_xfmr_topology}
\end{figure*}

After the sampling modules produce lower-resolution feature maps under the I/O topology, the spatially contiguous DMP-stripe layout must be converted into the window-major layout required by Swin Transformer execution. SAWSTP performs this conversion through the DMP token route. To support different Transformer execution strategies, SAWSTP represents the Transformer stage using a factorized topology $\mathrm{Topo}(m,n,s,t)$, where $(m,n)$ defines a two-dimensional WP grid for complete-window ownership, $s$ denotes the SP degree for intra-window sequence partitioning, and $t$ denotes the TP degree for sharding Transformer operators and intermediate activations. Under a selected topology, each Transformer rank is identified by a factorized coordinate $(i,j,k,l)$, where $(i,j)$ specifies the WP coordinate, $k$ identifies the intra-window SP coordinate, and $l$ identifies the TP coordinate. For clarity, $\mathrm{Topo}(m,n)$, $\mathrm{Topo}(m,n,s)$, and $\mathrm{Topo}(m,n,s,t)$ denote pure WP, WP+SP, and WP+SP+TP, respectively.

Fig.~\ref{fig_xfmr_topology} illustrates how the same DMP-stripe layout is transformed into the window-major layouts required by these Transformer topologies. Each complete attention window is represented by a $2\times2$ block with the same color, while the coordinate labels indicate the ranks responsible for the corresponding distributed window data. The gray boxes show the layouts directly produced by the DMP token route, and the layouts before the gray boxes conceptually show how complete windows are split into SP and TP sub-tiles. Under pure WP, each complete window is assigned to one WP rank, while the shift token route uses point-to-point (P2P) communication to redistribute tokens across WP ranks for shifted-window execution. When SP is enabled, each window is further partitioned along the intra-window sequence dimension, and Ulysses-based all-to-all communication reorganizes these sequence shards around window attention. 

The $\mathrm{Topo}(m,n,s,t)$ extends the WP+SP design of AERIS with FSDP and TP. AERIS-style WP+SP partitions window activations but does not shard Transformer parameters. AERIS primarily uses PP to reduce parameter-related memory. However, pipeline bubbles are difficult to amortize under the limited global batch sizes of high-resolution Earth forecasting models such as Wenhai. SAWSTP instead applies FSDP over the $\mathrm{DP}\times\mathrm{WP}\times\mathrm{SP}$ process group. For Transformer blocks with a large hidden dimension $d$, FSDP must all-gather the full parameters of each wrapped unit before its execution, which can cause OOM failures. SAWSTP therefore introduces Megatron-style TP~\cite{19megatron, 23-mlsys-megatron} within each WP--SP group. As shown in Fig.~\ref{fig_xfmr_topology}, TP further partitions the computation assigned to each SP rank. It uses all-gather to materialize the activations required for a SP rank and reduce-scatter to restore the TP-sharded activation states afterward. Therefore, for a fixed world size, TP further shards Transformer parameters while preserving the per-rank activation layout of AERIS-style WP+SP partitioning.

\subsection{Differentiable Token Routes}

SAWSTP couples a DMP-based I/O topology with a multi-level Transformer topology that integrates WP, SP, and TP. These topologies assign tensor ownership, layouts, and communication groups differently, making it nontrivial to ensure correct token routing in the forward pass and gradient propagation in the backward pass. To address this issue, TERRA indexes both the DMP-stripe and window-major layouts using the same global two-dimensional token grid. The corresponding conversion from DMP-stripe layout to window-major layout is implemented as the DMP token route.

After down-sampling, the Transformer-stage token grid has spatial size $H_T\times W_T$, where
\begin{equation}
H_T = \frac{\hat{H}}{p\alpha}, \qquad
W_T = \frac{\hat{W}}{p\alpha}.
\end{equation}
The corresponding attention-window grid contains $M_h\times M_w$ windows, where $M_h=H_T/w$ and $M_w=W_T/w$. Each Transformer-stage token on the token grid is identified by a global coordinate $(u,v)$. Its attention-window coordinate is $(\left\lfloor \frac{u}{w} \right\rfloor, \left\lfloor \frac{v}{w} \right\rfloor)$, and its within-window offset is determined by $(u\bmod w,v\bmod w)$. 

Under the DMP-stripe layout, $(u,v)$ identifies the source rank and local offset of a token in its spatial stripe. Given the selected $\mathrm{Topo}(m,n,s,t)$ topology and window-assignment policy, $(u,v)$ determines the owner of its complete attention window, its destination rank, and its local offset in the window-major tensor. Tokens with the same destination rank are gathered using \texttt{torch.index\_select} in PyTorch~\cite{19pytorch} and packed into contiguous send buffers. TERRA then exchanges the peer-wise buffers through batched P2P communication. Each destination rank restores received tokens at their destination positions to construct the required window-major tensor.

The shift token route uses the same coordinate-based mechanism. Window shifting changes the global coordinate $(u,v)$ of each token, from which TERRA determines its new destination rank and local offset. The corresponding tokens are gathered, communicated, and restored using the same indexed packing procedure. Both routes are implemented as autograd-aware communication primitives. Their backward passes apply the inverse mappings to return gradients to the corresponding source ranks and local offsets. This design allows TERRA to support different Transformer topologies and window-assignment policies and provides the basis for the window-assignment optimizations presented in Section~\ref{sec:cost_window_assignment}.

\section{Cost-Aware Window Assignment} 

\label{sec:cost_window_assignment}
\begin{figure}[!t]
\centering
\includegraphics[width=\columnwidth]{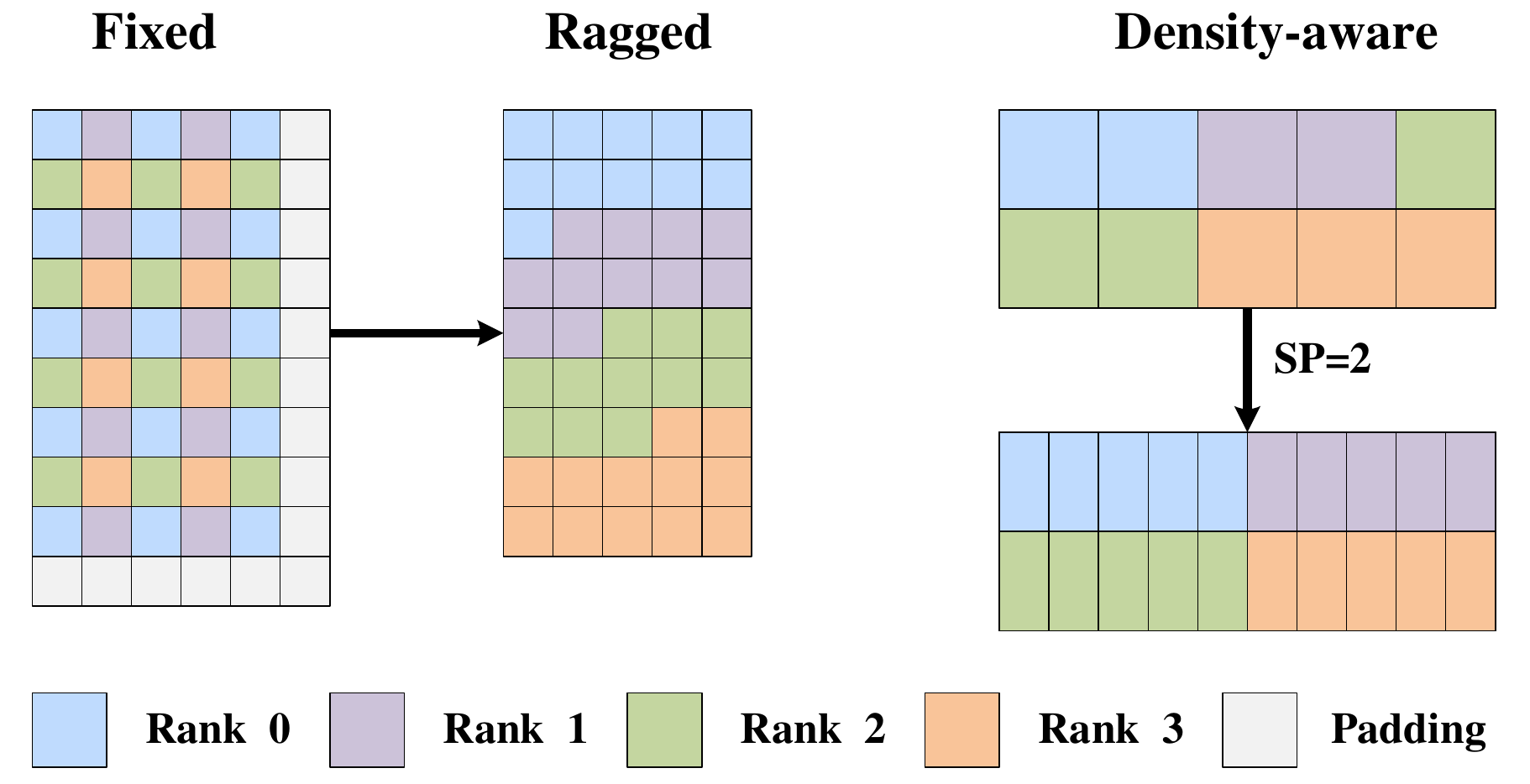}
\caption{Fixed, ragged, and density-aware window assignment.}
\label{fig_topology_padding}
\end{figure}

\subsection{Ragged Window Assignment}
The multi-level Transformer topology maps complete attention windows to ranks along the WP dimension. As shown in the left part of Fig.~\ref{fig_topology_padding}, the AERIS-style fixed assignment uses two-dimensional round-robin mapping to distribute the windows of an $M_h\times M_w$ grid across an $m\times n$ WP mesh. The illustrated example uses $M_h=9$, $M_w=5$, and $m=n=2$.

Fixed assignment provides uniform window ownership only when $M_h$ and $M_w$ are divisible by $m$ and $n$, respectively. Otherwise, the input spatial dimensions are padded to $\hat{H}\times\hat{W}$, where  
\begin{equation}
\hat{H}=
\left\lceil\frac{H}{p\alpha wm}\right\rceil p\alpha wm,\qquad
\hat{W}=
\left\lceil\frac{W}{p\alpha wn}\right\rceil p\alpha wn.
\end{equation}
This ensures that every WP rank receives the same number of complete windows. However, the additional padded tokens increase activation memory and computation. Moreover, two-dimensional round-robin ownership requires distributing attention windows to adjacent WP ranks, which brings substantial P2P communication during shifted-window execution.

To address this issue, TERRA instead adopts a ragged $\mathrm{Topo}(mn,1)$ assignment. As shown in the middle part of Fig.~\ref{fig_topology_padding}, it avoids padding introduced for equal window ownership and preserves row-major contiguous ownership. Let $M=M_hM_w$ be the total number of complete windows. After flattening the window grid in row-major order, the $M$ complete windows are indexed from $0$ to $M-1$. WP rank $r$, where $0\leq r<mn$, owns the contiguous window-index range $\left[L_r,R_r\right)$, where
\begin{equation}
L_r=\left\lfloor\frac{rM}{mn}\right\rfloor,\qquad
R_r=\left\lfloor\frac{(r+1)M}{mn}\right\rfloor.
\end{equation}

The ragged assignment is enabled by the DMP token route, which allows each WP rank to receive tokens from the DMP ranks for a different number of complete windows. Thus, every complete window is assigned exactly once, and the window counts of any two ranks differ by at most one. Moreover, only boundary windows whose shifted-window token fragments cross a WP-rank boundary require P2P communication, whereas windows in the interior of each rank's contiguous range remain local. This substantially reduces the communication volume of shifted-window execution.

\subsection{Density-Aware Window Assignment}
The ragged $\mathrm{Topo}(mn,1)$ assignment is effective when the window grid is dense enough to provide sufficient assignment granularity. As the window size $w$ increases, the number of complete windows decreases, and each complete window becomes a coarse assignment unit. The resulting ragged WP assignment can therefore become imbalanced. As shown in the right part of Fig.~\ref{fig_topology_padding}, assigning ten windows to four ranks with ragged $\mathrm{Topo}(4,1)$ gives two ranks three windows each and the other two ranks two windows each. When the WP group size exceeds the number of complete windows, pure-WP assignment can even leave some ranks with no assigned window.

SP can mitigate this load imbalance by partitioning each complete attention window across SP ranks. For example, setting $s=2$ changes the topology to $\mathrm{Topo}(2,1,2)$, and the resulting twenty window sub-tiles allow each of the four ranks to process five sub-tiles. However, SP introduces all-to-all communication around window attention. This tradeoff motivates TERRA's density-aware assignment, which extends the ragged WP assignment with SP when the performance benefit of improved assignment granularity covers the additional communication cost.

Table~\ref{tab:window_assignment_comm} summarizes the communication characteristics of the fixed baseline and the two TERRA assignments. AERIS-style fixed assignment incurs P2P communication during window shifting. Ragged assignment limits this overhead to rank-boundary windows. Density-aware assignment preserves this reduced P2P communication while additionally introducing SP all-to-all communication.

\begin{table}[!t]
\centering
\caption{Communication characteristics of window assignments.}
\label{tab:window_assignment_comm}
\footnotesize
\renewcommand{\arraystretch}{1.12}
\setlength{\tabcolsep}{4pt}
\begin{tabular}{l l}
\hline
\textbf{Strategy} & \textbf{Communication} \\ \hline
Fixed $\mathrm{Topo}(m,n)$ &
Two-dimensional WP P2P
\\
Ragged $\mathrm{Topo}(mn,1)$ &
Boundary-only P2P
\\
Density-aware $\mathrm{Topo}(mn/s,1,s)$ &
Boundary-only P2P + SP all-to-all
\\ \hline
\end{tabular}
\end{table}

Algorithm~\ref{alg:density_aware_assignment} describes an offline profiling-based strategy for density-aware window assignment. Given $mn$ ranks assigned to the WP and SP dimensions, TERRA enumerates the valid divisors $s$ of $mn$. For each $s$, it constructs the corresponding $\mathrm{Topo}(mn/s,1,s)$ candidate, profiles the candidate for several iterations, and selects the topology with the lowest average training-step time. The selected topology is then used throughout the subsequent training steps. For high window density, $s=1$ is typically selected because ragged WP already provides sufficient load balance without introducing all-to-all communication. For low window density, a larger $s$ is selected only when its load-balancing benefit outweighs the additional SP communication overhead.

\begin{algorithm}[t]
\caption{Density-aware window assignment}
\label{alg:density_aware_assignment}
\begin{algorithmic}[1]
\STATE \textbf{Input:} rank count $mn$ for WP and SP
\STATE \textbf{Output:} selected topology
\STATE $\mathit{bestTime}\leftarrow+\infty$
\FOR{each valid divisor $s$ of $mn$}
    \STATE $\mathit{candidateTopology}\leftarrow
    \mathrm{Topo}(mn/s,1,s)$
    \STATE $\mathit{candidateTime}\leftarrow
    \mathrm{Profile}(\mathit{candidateTopology})$
    \IF{$\mathit{candidateTime}<\mathit{bestTime}$}
        \STATE $\mathit{bestTime}\leftarrow
        \mathit{candidateTime}$
        \STATE $\mathit{bestTopology}\leftarrow
        \mathit{candidateTopology}$
    \ENDIF
\ENDFOR
\STATE \textbf{return} $\mathit{bestTopology}$
\end{algorithmic}
\end{algorithm}

\section{Memory Orchestration}\label{sec:memory_orchestration}

\begin{figure*}[!t]
\centering
\subfloat[$D_{0}$, $U_{0}$, $t_{\mathrm{seg}}=1$.]{
  \includegraphics[width=0.23\textwidth]{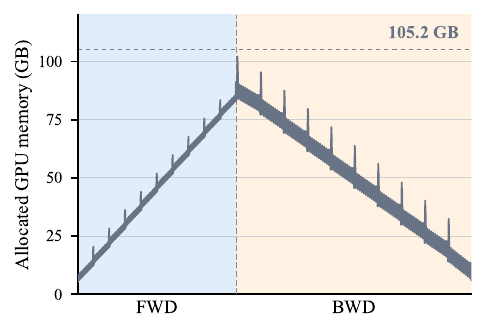}
  \label{fig:orchestra_timeline_t1}
}
\hfil
\subfloat[$D_{0}$, $U_{0}$, $t_{\mathrm{seg}}=3$.]{
  \includegraphics[width=0.23\textwidth]{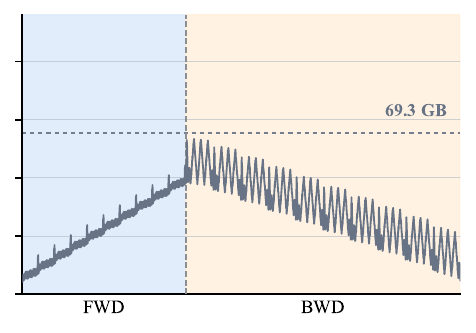}
  \label{fig:orchestra_timeline_t3}
}
\hfil
\subfloat[Checkpoint planning.]{
  \includegraphics[width=0.23\textwidth]{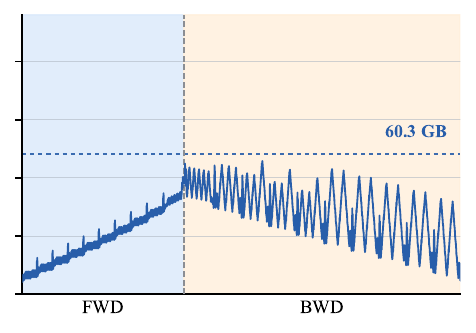}
  \label{fig:orchestra_timeline_dp}
}
\hfil
\subfloat[Budget-constrained offloading.]{
  \includegraphics[width=0.23\textwidth]{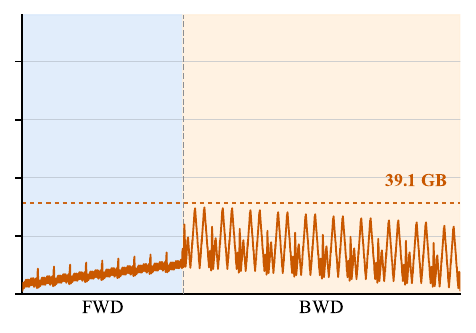}
  \label{fig:orchestra_timeline_offload}
}
\caption{Allocated-memory timelines for ten-lead finetuning under different policies.}
\label{fig:orchestra_timeline}
\end{figure*}

\subsection{Activation Bottleneck in Finetuning}

SAWSTP reduces the per-rank memory footprint of hierarchical Swin Transformer training. However, long-lead finetuning remains dominated by activation memory, particularly with small patch sizes. Fig.~\ref{fig:orchestra_timeline} shows the allocated GPU-memory timelines during the forward (FWD) and backward (BWD) passes of ten-lead rollout finetuning. During FWD, checkpoint boundaries generated at each lead remain resident until their corresponding BWD computations and therefore accumulate across the rollout. During BWD, recomputation of a sampling or Transformer checkpoint unit produces a short-lived local peak.

Although numerous memory-optimization techniques based on activation recomputation~\cite{16-chen-ckpt,16bptt,20dtr,23rockmate} and activation offloading~\cite{16vdnn,19vdnn++,21-nips-ckpt,24delta} have been proposed, it remains nontrivial to jointly determine recomputation granularity and offloading decisions for hierarchical rollout finetuning under SAWSTP. Sampling and Transformer checkpoint units exhibit different checkpoint-boundary tensor sizes and recomputation peaks, while high-resolution input and label tensors further introduce a non-negligible baseline memory footprint. These characteristics require an architecture-aware policy that models rollout-wide tensor lifetimes and satisfies a target GPU-memory budget.

TERRA addresses this challenge through Memory Orchestration (MO), a profile-based two-stage policy. MO first constructs an architecture-aware checkpoint-only plan by modeling rollout-wide boundary accumulation and local recomputation peaks. When checkpointing alone cannot satisfy a target GPU-memory budget, MO combines input buffering with selective boundary offloading to meet the budget.

\subsection{Profile-Based Checkpoint Planning}
The first stage of MO constructs a checkpoint-only policy from a block-wise runtime memory profile under the selected SAWSTP configuration. Inspired by the runtime memory tracer in PatrickStar~\cite{22patrickstar}, TERRA records the profile at runtime and visualizes the allocated GPU-memory timeline. It uses \texttt{torch.cuda.memory\_allocated} for the allocated-memory timeline and \texttt{torch.cuda.max\_memory\_allocated} for peak memory. As shown in Fig.~\ref{fig:orchestra_timeline}, the $p=2$, $\mathrm{DMP}=4$, and $\mathrm{DP}=2$ configuration exhibits distinct forward and backward memory patterns under different checkpointing policies.

MO distinguishes sampling and Transformer checkpoint units because they exhibit different boundary activation sizes and recomputation peaks. As shown in the lower-left part of Fig.~\ref{fig_parallel_framework}, each sampling module contains several Conv-based residual blocks. For sampling modules, MO considers two checkpoint granularities. At the coarse granularity, $D_{0}$ treats patch embedding and the subsequent down-sampling module as one checkpoint unit, whereas $U_{0}$ combines the up-sampling module with patch recovery. At the fine granularity, $D_{1}$ combines patch embedding with the first stride-two convolution and checkpoints each subsequent residual block independently. $U_{1}$ applies the analogous fine-grained partitioning to the up-sampling module and patch recovery.

The Transformer stage contains $L=10$ Swin Transformer blocks. MO profiles checkpoint units containing $t_{\mathrm{seg}}\in\{1,2,3,4,5,10\}$ consecutive blocks. Coarser units retain fewer checkpoint boundaries across the rollout but produce larger peak memory during recomputation, whereas finer units have the opposite tradeoff. Figs.~\ref{fig:orchestra_timeline}
\subref{fig:orchestra_timeline_t1} and
\ref{fig:orchestra_timeline}\subref{fig:orchestra_timeline_t3} illustrate this tradeoff for $D_{0}/U_{0}$ with $t_{\mathrm{seg}}=1$ and $t_{\mathrm{seg}}=3$, respectively. In particular, the coarse $D_{0}/U_{0}$ units can substantially reduce boundary memory because their internal boundaries correspond to high-resolution feature maps.

For each checkpoint unit $i$, MO profiles its boundary memory $E_i$ and backward peak $P_i$. A checkpoint-only policy $\pi$ selects the sampling granularity and Transformer segment length at each rollout lead. MO predicts the resulting peak memory as
\begin{equation}
M(\pi)=M_{\mathrm{base}}+
\max_i\left\{A_i(\pi)+P_i(\pi)\right\},
\label{eq:orchestra_memory_model}
\end{equation}
where $A_i(\pi)$ denotes the cumulative memory of checkpoint boundaries that remain resident immediately before the backward execution of unit $i$. The model accumulates boundaries that coexist across rollout leads and takes the maximum over the transient peaks created by recomputation. $M_{\mathrm{base}}$ denotes the profiled fixed memory footprint, including input and label tensors and other persistent runtime allocations. MO minimizes Eq.~(\ref{eq:orchestra_memory_model}) using a dynamic-programming method inspired by~\cite{16bptt}. As illustrated in Fig.~\ref{fig:orchestra_timeline}\subref{fig:orchestra_timeline_dp}, the dynamic-programming method selects coarser checkpoint units for early leads and finer-grained units for later leads in this configuration, reducing retained boundary memory and recomputation peaks, respectively.

\subsection{Budget-Constrained Offloading}

When the checkpoint-only policy $\pi$ cannot satisfy a target memory budget $M_b$, MO constructs an offloading-based policy $\pi_o$ using the profiled boundary memories $E_i$ and backward peaks $P_i$. As shown in Algorithm~\ref{alg:budget_repair}, MO employs a lightweight profile-guided greedy algorithm to satisfy the memory constraint without solving a globally optimal tensor-level offloading schedule. MO first considers $D_{0}/U_{0}$ and selects the largest $t_{\mathrm{seg}}$ whose profile-predicted peak memory does not exceed $M_b$. The selected sampling granularity and Transformer segment template are then applied to all rollout leads. MO falls back to $D_{1}/U_{1}$ only when no $D_{0}/U_{0}$ configuration satisfies the memory constraint.

MO then constructs $\pi_o$ over the $K$ rollout leads. For each lead $k$, it appends the selected configuration to $\pi_o$ and inserts its down-sampling, Transformer, and up-sampling boundaries into a queue $\mathcal{Q}$ in forward creation order. $M_k(\pi_o)$ denotes the profile-predicted peak obtained by replaying the backward events of lead $k$ under the current offloading policy. It includes resident boundary memory, backward peak increments, and the boundary memory restored for
offloaded checkpoint units. If $M_k(\pi_o)$ exceeds $M_b$, MO removes the earliest boundary from the queue and marks it for offloading. Earlier boundaries normally have longer forward-to-backward lifetimes, making them suitable candidates for asynchronous offloading. The procedure continues until the memory constraint is satisfied.

Fig.~\ref{fig:orchestra_timeline}\subref{fig:orchestra_timeline_offload} illustrates the resulting offloading-based memory timeline under a $40$~GB budget. Compared with the checkpoint-only policy in Fig.~\ref{fig:orchestra_timeline}\subref{fig:orchestra_timeline_dp}, boundary offloading reduces the measured peak from $60.3$~GB to $39.1$~GB, which satisfies the memory constraint.

\begin{algorithm}[t]
\caption{Profile-guided boundary activation offloading}
\label{alg:budget_repair}
\begin{algorithmic}[1]
\STATE \textbf{Input:} profiled $E_i$, $P_i$, and $M_{\mathrm{base}}$, rollout length $K$, and memory budget $M_b$
\STATE \textbf{Output:} offloading-based policy $\pi_o$

\STATE A configuration $(D,t_{\mathrm{seg}},U)$ is feasible if $M_{\mathrm{base}}+E_i+P_i\leq M_b$ for every checkpoint unit $i$
\STATE Select the coarsest sampling granularity and the largest feasible $t_{\mathrm{seg}}$ for $(D,t_{\mathrm{seg}},U)$
\STATE $\pi_o\leftarrow\emptyset$ and $\mathcal{Q}\leftarrow\emptyset$

\FOR{$k=1$ to $K$}
    \STATE Append $(D,t_{\mathrm{seg}},U)$ for lead $k$ to $\pi_o$
    \STATE Append its checkpoint boundaries to $\mathcal{Q}$ in
    forward creation order
    \STATE Compute the predicted peak $M_k(\pi_o)$

    \WHILE{$M_k(\pi_o)>M_b$}
        \STATE Remove the earliest boundary $i$ from $\mathcal{Q}$
        \STATE Mark boundary $i$ for offloading in $\pi_o$
        \STATE Recompute the predicted peak $M_k(\pi_o)$
    \ENDWHILE
\ENDFOR

\STATE \textbf{return} $\pi_o$
\end{algorithmic}
\end{algorithm}

In the implementation, TERRA applies a boundary-offload wrapper to each checkpoint unit selected by MO. The wrapper extends the \texttt{torch.utils.checkpoint} interface with \texttt{torch.autograd.graph.saved\_tensors\_hooks}, which capture the input tensors saved at each checkpoint boundary. The pack hook offloads checkpoint boundaries to pinned host buffers through asynchronous device-to-host (D2H) transfers, while the unpack hook restores them through host-to-device (H2D) transfers before backward recomputation. This design requires minimal changes to the training code and is compatible with both FSDP and SAWSTP.

\subsection{Input Buffering}

Even with activation checkpointing and offloading, high-resolution input and label data consume substantial GPU memory and constitute a major component of $M_{\mathrm{base}}$. This cost is particularly large at small DMP degrees, where each rank owns a larger spatial shard. For Wenhai, one FP16 field of shape $93\times 2041\times 4320$ occupies approximately $1.53$~GB. A ten-lead rollout requires one input field and ten label fields for loss computation. Therefore, simultaneously transferring all these fields via H2D incurs both substantial transfer overhead and a large GPU-memory footprint.

To address this issue, TERRA employs a lead-wise input and label prefetching pipeline with pinned host memory and non-blocking H2D transfers. Rather than transferring the entire rollout sequence to the GPU at once, the pipeline transfers tensors on demand and prefetches data for upcoming leads during rollout execution. Its input-prefetch stream is separate from the activation-offload stream, allowing H2D prefetching and checkpoint-boundary D2H offloading to progress independently when resources permit. Together, the input buffering method bounds the GPU-side lifetime of input and label tensors and hides part of the H2D latency behind rollout computation, thereby reducing GPU-memory consumption and end-to-end step time.

\section{Evaluation}

\subsection{Experiment Setup}
\label{sec:experimental_setup}

\subsubsection{Hardware and Software}
We conduct all experiments on a homogeneous cluster comprising up to 12 nodes and 96 NVIDIA H200 GPUs. Each GPU provides 141\,GB of HBM3e memory and a peak dense FP16 throughput of 989\,TFLOPS. Table~\ref{tab:experimental_platform} summarizes the hardware environment. The software stack comprises PyTorch 2.4.0, CUDA 12.4, NCCL 2.20.5, FlashAttention 2.7.3~\cite{22FA}, and NVIDIA driver 570.133.20.

\begin{table}[!t]
\centering
\caption{Experimental platform.}
\label{tab:experimental_platform}
\scriptsize
\renewcommand{\arraystretch}{1.08}
\setlength{\tabcolsep}{3.5pt}
\begin{tabular}{c|p{0.67\columnwidth}}
\hline
\textbf{Component} & \textbf{Configuration} \\ \hline
Cluster scale & Up to 12 nodes and 96 GPUs \\
GPU & $8\times$ NVIDIA H200 GPUs per node \\
CPU & $2\times$ Intel Xeon Platinum 8558, 96 cores per node \\
Host memory & 1.8\,TB per node \\
Intra-node interconnect & NVLink \\
Inter-node interconnect & Mellanox RDMA network \\

\hline
\end{tabular}
\end{table}

\subsubsection{Workload and Models}

\begin{table}[!t]
\centering
\caption{Model configurations.}
\label{tab:scaling_models}
\scriptsize
\renewcommand{\arraystretch}{1.08}
\setlength{\tabcolsep}{3.5pt}
\begin{tabular}{c|ccccccc}
\hline
\textbf{Model} &
$L$ &
$d$ &
$N_{\mathrm{h}}$ &
$p$ &
$\alpha$ &
$w$ &
\textbf{Number of Parameters} \\ \hline
Small  & 10 & 1024 & 8  & 4 & 2 & 8  & 0.18B \\
Medium & 10 & 4096 & 32 & 4 & 2 & 8  & 2.85B \\
Large  & 10 & 8192 & 64 & 4 & 2 & 32 & 11.4B \\ \hline
\end{tabular}
\end{table}

We evaluate TERRA on the GLORYS-based Wenhai workload, in which each training sample contains 93 physical variables on a $2041\times4320$ global ocean grid. Table~\ref{tab:scaling_models} summarizes the small, medium, and large hierarchical Wenhai model configurations used in our experiments. Models are trained using FP16 mixed precision and mean absolute error (MAE) loss. Unless otherwise specified, performance measurements use a micro-batch size of $B=1$, while end-to-end training and forecasting-accuracy experiments use a global batch size of 16. Transformer checkpoint units use $t_{\mathrm{seg}}=1$, and the down- and up-sampling modules use $D_{1}/U_{1}$ checkpointing.

For the small model, we apply a single FSDP wrapper to the entire model. For the medium and large models, we wrap each Transformer block and sampling module independently with FSDP. FlashAttention is enabled for configurations with window size $w\geq32$.

\subsubsection{Baselines and Metrics}
We evaluate the scalability of TERRA through strong- and weak-scaling experiments on hierarchical Wenhai models. Unless otherwise specified, the reported step time includes the forward, backward, and optimizer-update phases but excludes the loading and H2D transfer of input and label tensors. For parameter-capacity scaling, we compare an AERIS-style WP+FSDP baseline ($\mathrm{TP}=1$) with TERRA configurations that combine FSDP and TP. For window assignment, we compare TERRA's ragged $\mathrm{Topo}(mn,1)$ and density-aware $\mathrm{Topo}(mn/s,1,s)$ assignments with fixed AERIS-style WP and WP+SP topologies. For MO, we evaluate its effectiveness in reducing GPU memory consumption during rollout finetuning under different SAWSTP configurations.

For forecasting accuracy, we compare models with different patch sizes and sampling configurations, as well as models finetuned with different rollout lengths. These comparisons quantify the forecasting benefits enabled by TERRA through finer patch granularity, hierarchical sampling, and longer rollout finetuning. We report the mean root mean square error (RMSE) over all predicted variables, where lower values indicate more accurate forecasts.

\subsection{Convergence Validation}

\begin{figure}[!t]
\centering
\subfloat[Steps 0--100.]{
  \includegraphics[width=0.470\columnwidth]{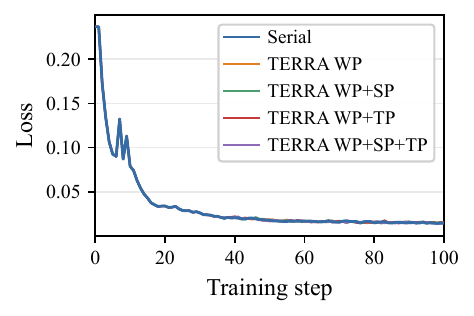}
  \label{fig:loss_curves_early}
}
\hfill
\subfloat[Steps 100--1000.]{
  \includegraphics[width=0.470\columnwidth]{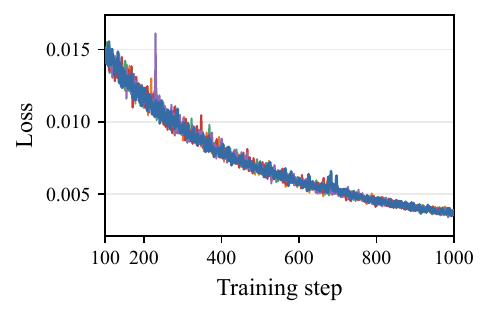}
  \label{fig:loss_curves_late}
}
\caption{Training-loss consistency between serial execution and TERRA parallel configurations.}
\label{fig:loss_curves}
\end{figure}

We first validate the correctness of TERRA's parallel training framework. Fig.~\ref{fig:loss_curves} compares the training-loss curve obtained from serial execution of the small model in Table~\ref{tab:scaling_models} with the corresponding curves obtained from TERRA's parallel configurations. The serial execution uses FSDP alone and does not employ SAWSTP. For a fair comparison, all runs use the same padded input resolution of $2304\times4352$ and $\mathrm{DP}=2$.

The TERRA configurations use eight ranks per SAWSTP group and vary the Transformer topology across WP, SP, and TP. The WP, WP+SP, WP+TP, and WP+SP+TP configurations use ragged $\mathrm{Topo}(8,1,1,1)$, $\mathrm{Topo}(4,1,2,1)$, $\mathrm{Topo}(4,1,1,2)$, and $\mathrm{Topo}(2,1,2,2)$, respectively. As shown in Fig.~\ref{fig:loss_curves}, all TERRA configurations exhibit convergence behavior comparable to that of serial execution over the first $1000$ training steps. This agreement validates gradient propagation through parallel sampling, token routing, and the evaluated combinations of WP, SP, and TP.

\subsection{Scalability}

\begin{figure*}[!t]
\centering
\subfloat[Strong scaling step time and efficiency.]{
  \includegraphics[width=0.45\textwidth]{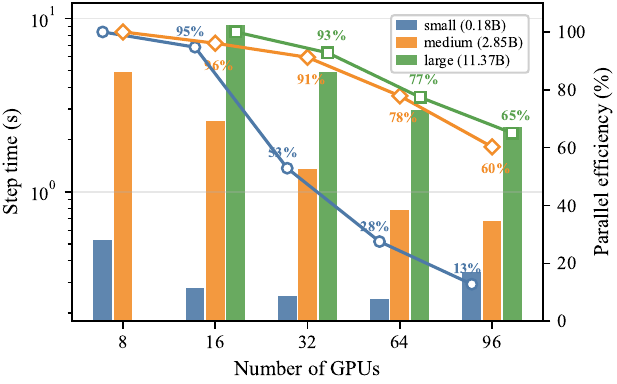}
  \label{fig:strong_scaling_time_eff}
}
\hfil
\subfloat[Peak memory under strong scaling.]{
  \includegraphics[width=0.45\textwidth]{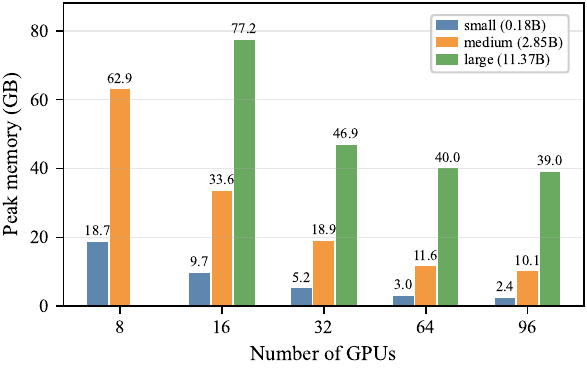}
  \label{fig:strong_scaling_memory}
}
\caption{Strong scaling performance.}
\label{fig:strong_scaling}
\end{figure*}

\begin{figure*}[!t]
\centering
\subfloat[Weak scaling throughput.]{
  \includegraphics[width=0.45\textwidth]{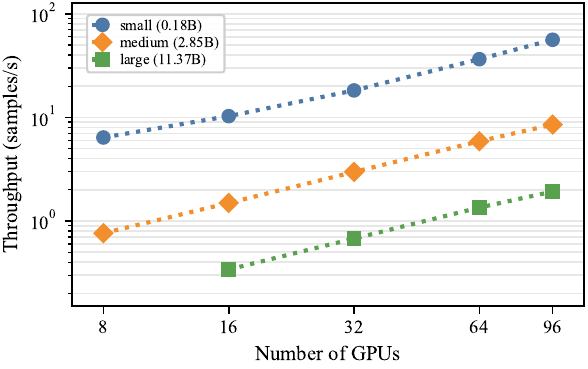}
  \label{fig:weak_scaling_throughput}
}
\hfil
\subfloat[Weak scaling sustained performance.]{
  \includegraphics[width=0.45\textwidth]{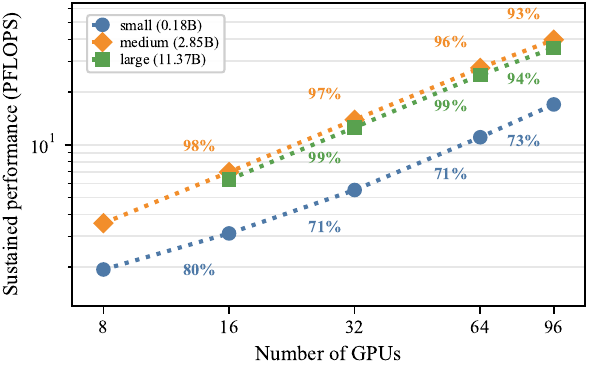}
  \label{fig:weak_scaling_performance}
}
\caption{Weak scaling performance.}
\label{fig:weak_scaling}
\end{figure*}

We evaluate the strong- and weak-scaling performance of TERRA using the Wenhai models in Table~\ref{tab:scaling_models}. To avoid OOM errors, measurements for the large model start at 16 GPUs. For strong scaling, we fix $\mathrm{DP}=4$ and increase the DMP degree with the number of GPUs. For weak scaling, we fix $\mathrm{DMP}=8$ for the small and medium models and $\mathrm{DMP}=16$ for the large model, while increasing the DP degree with the number of GPUs. For all configurations, we set $s=1$ and use the ragged $\mathrm{Topo}(mn,1)$ assignment.

Fig.~\ref{fig:strong_scaling} shows the strong-scaling results. Increasing the DMP degree reduces peak allocated GPU memory for all three models and consistently decreases the step time of the medium and large models. These results show that SAWSTP distributes the computation and activations of both sampling modules and Transformer blocks across multiple GPUs. The large model achieves a $3.9\times$ speedup from 16 to 96 GPUs, corresponding to $65.0\%$ strong-scaling efficiency. In contrast, the strong scalability of the small model is limited because the DMP stripe and the number of attention windows assigned to each rank become too small at high GPU counts, while communication and synchronization overheads become increasingly significant. Increasing the micro-batch size could provide more per-rank computation to amortize these overheads and thereby improve throughput.

Fig.~\ref{fig:strong_scaling}\subref{fig:strong_scaling_memory} further shows that activation and parameter-related memory scale differently. As the SAWSTP group size increases, DMP distributes high-resolution sampling activations, while WP distributes Transformer activations. The larger FSDP sharding group also reduces per-rank parameter-related memory. Consequently, the peak memory of the small and medium models continues to decrease as more GPUs are used. However, peak memory remains around $40$~GB at 64 and 96 GPUs for the large model. In this case, full-parameter materialization during FSDP all-gather accounts for an increasing fraction of the per-rank peak. This observation motivates combining TP with FSDP to scale larger hierarchical Swin Transformer models.

Fig.~\ref{fig:weak_scaling} shows the weak-scaling results. The medium model increases aggregate throughput from $0.76$ to $8.5$ samples/s between 8 and 96 GPUs and achieves $39.76$~PFLOPS, corresponding to $93.0\%$ weak-scaling efficiency. The large model scales from $0.34$ samples/s on 16 GPUs to $1.93$ samples/s on 96 GPUs, achieving $94.1\%$ weak-scaling efficiency. The small model increases throughput from $6.4$ to $56.1$ samples/s but reaches a lower efficiency of $73.0\%$ at 96 GPUs because its smaller per-rank workload makes FSDP and SAWSTP communication more prominent.

\begin{figure}[!t]
\centering
\includegraphics[width=\columnwidth]{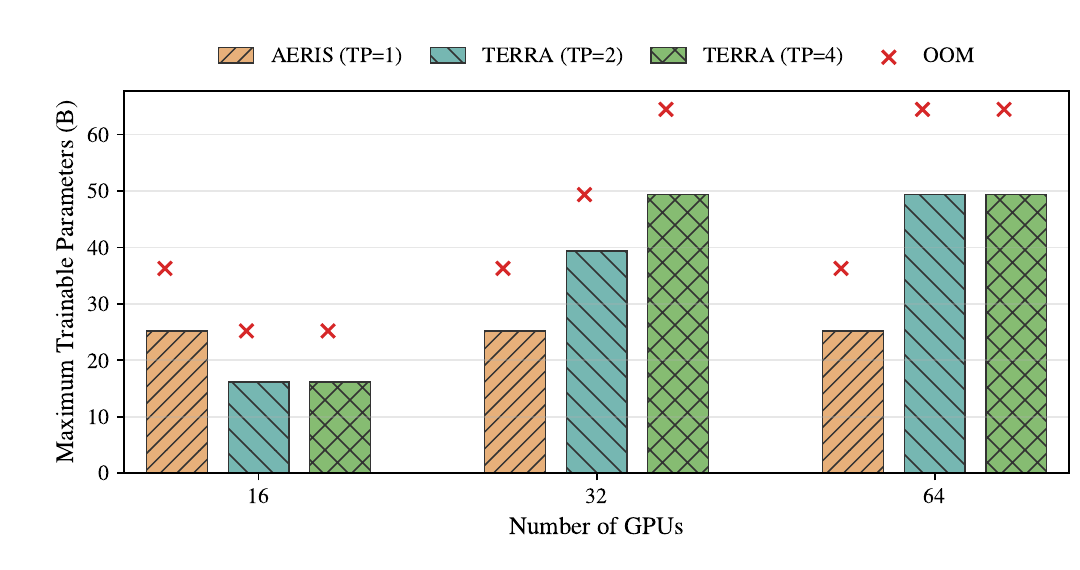}
\caption{Largest trainable model size for the AERIS-style baseline and TERRA configurations with TP.}
\label{fig:tp_vs_fsdp}
\end{figure}

Fig.~\ref{fig:tp_vs_fsdp} evaluates how TP complements WP and FSDP for parameter scaling. To focus the evaluation on parameter scaling in the Transformer blocks, we use a Swin Transformer without sampling modules and fix $L=20$, $p=4$, and $w=8$. We set $\mathrm{DP}=2$ and decompose the remaining ranks into WP and TP dimensions. The AERIS-style baseline retains AERIS's window-parallel execution and uses FSDP in the same runtime with $\mathrm{TP}=1$. The TERRA configurations use the same setup while additionally enabling $\mathrm{TP}=2$ or $4$. For each configuration, we increase the hidden dimension $d$ until the next candidate model runs out of memory and report the largest tested trainable model.

At 16 GPUs, the AERIS-style $\mathrm{TP}=1$ baseline supports 25.2B parameters, whereas the TERRA configurations with $\mathrm{TP}=2$ and $\mathrm{TP}=4$ support 16.1B parameters. At this scale, increasing the TP degree reduces the WP degree and causes each WP rank to process more windows. The activation all-gather operations required by TP further limit the memory savings from parameter partitioning at this scale. At 32 GPUs, the TERRA configurations with $\mathrm{TP}=2$ and $\mathrm{TP}=4$ increase the largest tested model size from 25.2B to 39.3B and 49.3B parameters, respectively. At 64 GPUs, both TERRA configurations support 49.3B parameters, while the AERIS-style baseline remains limited to 25.2B. By combining TP with FSDP, SAWSTP supports up to $2.0\times$ as many parameters as the $\mathrm{TP}=1$ baseline.

\subsection{Window Assignment Performance}

\begin{figure}[!t]
\centering
\subfloat[Training-step time on 16 GPUs.]{
  \includegraphics[width=0.45\textwidth]{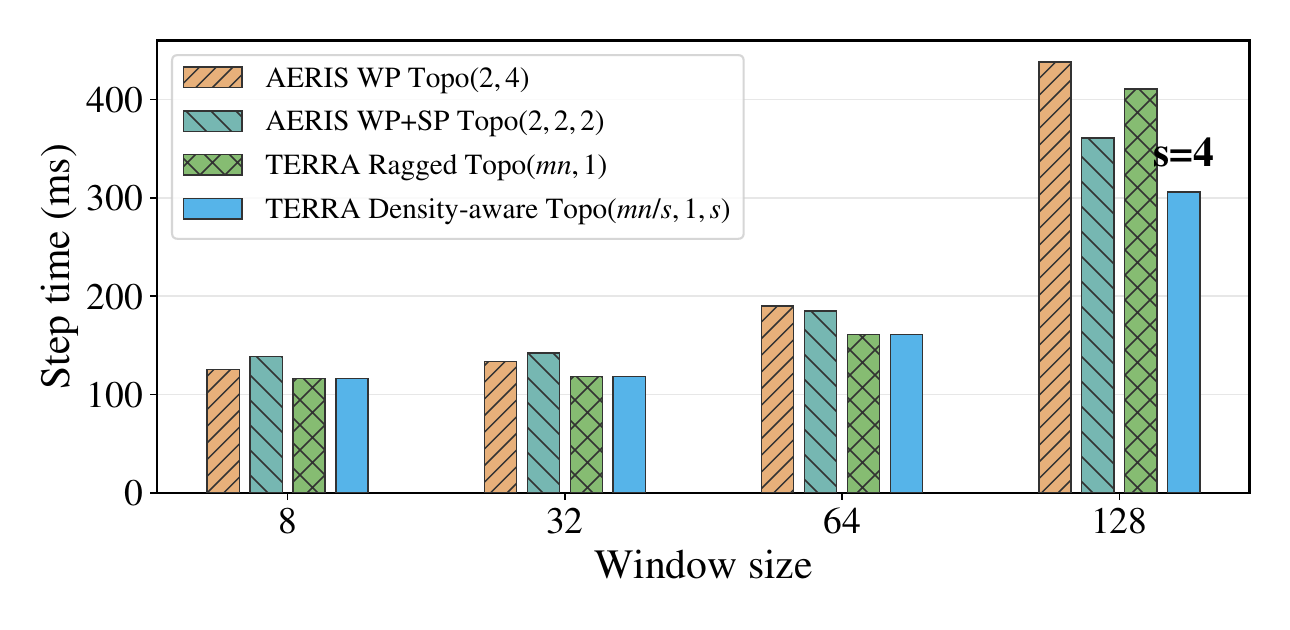}
  \label{fig:topology_policy_gpu16}
}

\vspace{0.6em}

\subfloat[Training-step time on 32 GPUs.]{
  \includegraphics[width=0.45\textwidth]{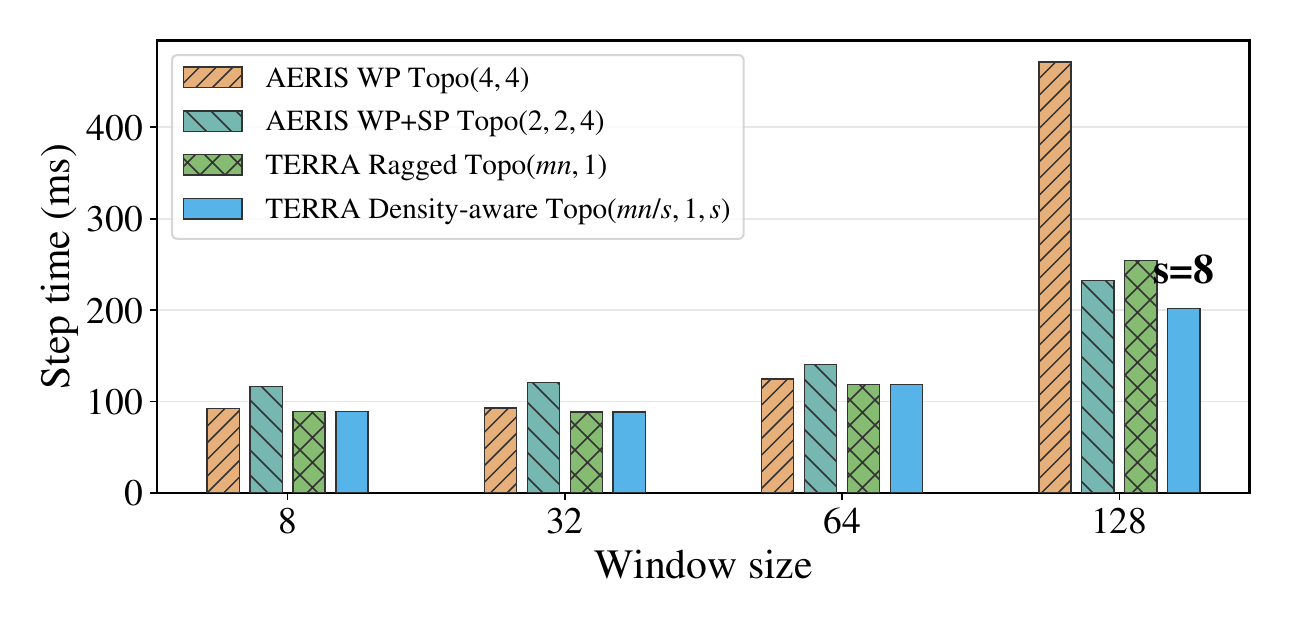}
  \label{fig:topology_policy_gpu32}
}
\caption{Training-step time under fixed AERIS-style topologies and TERRA's ragged window assignments.}
\label{fig:exp_topology_policy}
\end{figure}

Fig.~\ref{fig:exp_topology_policy} compares the window-assignment policies for the small model on 16 and 32 GPUs. Both configurations use $\mathrm{DP}=2$, with $\mathrm{DMP}=8$ on 16 GPUs and $\mathrm{DMP}=16$ on 32 GPUs. The AERIS-style baselines use fixed WP and WP+SP topologies. For TERRA, we report the ragged $\mathrm{Topo}(mn,1)$ topology without SP and the fastest measured policy among density-aware $\mathrm{Topo}(mn/s,1,s)$ candidates for each window size.

Overall, TERRA achieves a $1.03\times$--$1.18\times$ speedup over the faster AERIS-style baseline. For dense window grids ($w\leq64$), TERRA selects the ragged $\mathrm{Topo}(8,1)$ topology on 16 GPUs and $\mathrm{Topo}(16,1)$ on 32 GPUs. In this regime, complete-window WP provides sufficient parallelism, while introducing SP incurs unnecessary intra-window all-to-all communication. The ragged assignments also avoid padding introduced solely to maintain uniform window ownership when the window grid is incompatible with a fixed topology. Consequently, TERRA achieves $1.03\times$--$1.14\times$ speedups over the faster fixed baseline on 16 GPUs and $1.04\times$--$1.10\times$ speedups on 32 GPUs.

For the sparse-window case at $w=128$, TERRA selects the SP-enabled ragged topologies $\mathrm{Topo}(2,1,4)$ and $\mathrm{Topo}(2,1,8)$ on 16 and 32 GPUs, respectively. These policies reduce the step time to $306.4$~ms and $200.0$~ms, corresponding to $1.18\times$ and $1.14\times$ speedups over the faster fixed baseline. The worst case occurs for fixed WP on 32 GPUs. To maintain uniform window ownership, the fixed $\mathrm{Topo}(4,4)$ baseline pads the minimally padded $2\times5$ window grid to a $4\times8$ grid containing 32 windows. This increases the padding overhead to $280.6\%$, compared with $18.9\%$ under the selected ragged $\mathrm{Topo}(2,1,8)$ policy. Therefore, the selected policy achieves $2.36\times$ and $1.14\times$ speedups over the fixed $\mathrm{Topo}(4,4)$ WP-only and $\mathrm{Topo}(2,2,4)$ WP+SP baselines, respectively.

\subsection{Memory Orchestration Performance}

\begin{figure}[!t]
  \centering
  \includegraphics[width=0.45\textwidth]{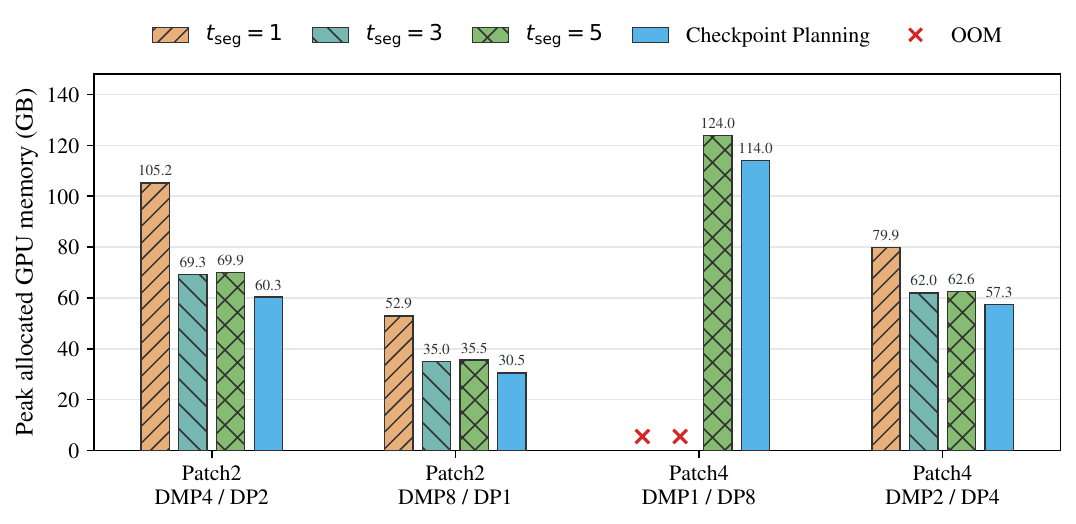}
  \caption{Peak allocated GPU memory of checkpoint-only policies for ten-lead
  finetuning.}
  \label{fig:orchestra_multiconfig}
\end{figure}

Fig.~\ref{fig:orchestra_multiconfig} evaluates the checkpoint-planning stage of MO for ten-lead finetuning on a single H200 node, with activation offloading and input buffering disabled. We use the small model and evaluate two SAWSTP configurations for each patch size. The $p=2$ experiments use $\mathrm{DMP}=4$ with $\mathrm{DP}=2$ and $\mathrm{DMP}=8$ with $\mathrm{DP}=1$, whereas the $p=4$ experiments use $\mathrm{DMP}=1$ with $\mathrm{DP}=8$ and $\mathrm{DMP}=2$ with $\mathrm{DP}=4$. We compare the resulting lead-dependent plan with uniform policies that use $D_0/U_0$ and a fixed $t_{\mathrm{seg}}\in\{1,3,5\}$ for all leads. Across the four SAWSTP configurations, checkpoint planning reduces peak allocated GPU memory by $7.6\%$--$13.0\%$ relative to the best feasible uniform policy. For $p=4$, $\mathrm{DMP}=1$, and $\mathrm{DP}=8$, the uniform $t_{\mathrm{seg}}=1$ and $t_{\mathrm{seg}}=3$ policies run out of memory, while $t_{\mathrm{seg}}=5$ reaches $124.0$~GB. Checkpoint planning remains feasible and reduces the peak to $114.0$~GB. These results show that a uniform checkpoint granularity cannot consistently balance rollout-wide boundary accumulation against local recomputation peaks.

\begin{figure}[!t]
  \centering
  \includegraphics[width=\columnwidth]{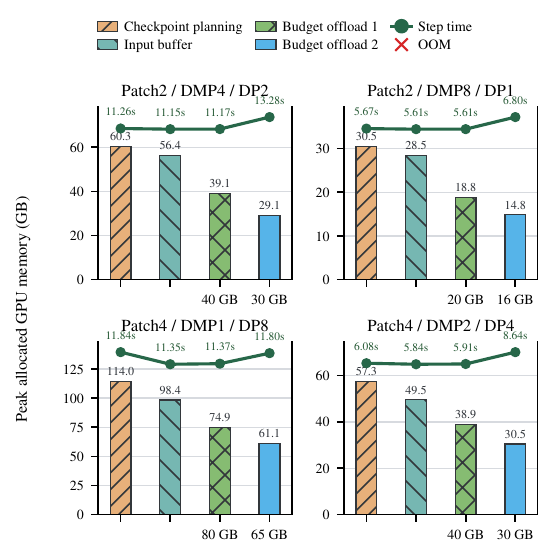}
  \caption{Peak allocated GPU memory and step time for input buffering and budget-constrained offloading.}
  \label{fig:orchestra_offload}
\end{figure}

Fig.~\ref{fig:orchestra_offload} evaluates input buffering and budget-constrained boundary offloading using the same four ten-lead configurations as Fig.~\ref{fig:orchestra_multiconfig}. The reported step time includes input and label H2D transfers as well as activation D2H and H2D transfers when offloading is enabled. During rollout execution, TERRA asynchronously prefetches the tensors required by the next lead and releases tensors that are no longer needed as execution advances. Across the four configurations, input buffering reduces peak allocated GPU memory by $6.4\%$--$13.7\%$ and step time by $1.0\%$--$4.1\%$. These results show that bounding the GPU-side lifetime of input and label tensors reduces memory consumption while overlapping part of their H2D transfers with computation.

MO enables activation offloading when the checkpoint-only policy exceeds the target budget $M_b$. Overall, relative to the corresponding checkpoint-only policies, the combination of input buffering and activation offloading reduces peak allocated GPU memory by $32.2\%$--$51.8\%$, with at most $20.0\%$ step-time overhead across the budget-feasible configurations. Under the primary budgets of $40$, $20$, $80$, and $40$~GB, all four configurations satisfy their targets, reducing peak memory by $32.2\%$--$38.2\%$. When the budgets are tightened to $30$, $16$, $65$, and $30$~GB, the first three configurations remain feasible and reduce peak memory by $46.4\%$--$51.8\%$, with at most $20.0\%$ step-time overhead. The $p=4$, $\mathrm{DMP}=2$, and $\mathrm{DP}=4$ configuration reaches a measured peak of $30.5$~GB and therefore slightly exceeds its $30$~GB budget. This difference results from a small prediction error in the profile-based memory model. In practice, OOM can be avoided by setting a slightly lower planning budget.

Table~\ref{tab:offload-volume} provides a stage-wise breakdown of the per-GPU activation offload volume per training step. When a tighter budget cannot be satisfied using the coarse $D_0/U_0$ units, MO switches to $D_1/U_1$, exposing additional sampling boundaries for offloading. For example, for $p=2$, $\mathrm{DMP}=4$, and $\mathrm{DP}=2$, tightening $M_b$ from $40$~GB to $30$~GB increases the total offload volume from $26.4$~GB to $102.4$~GB per GPU per step. This increase is dominated by sampling boundaries, whose offload volume grows from $9.7$~GB to $72.5$~GB, while the Transformer-boundary volume grows from $16.7$~GB to $29.9$~GB. The resulting increase in data transfer is consistent with the higher step times in Fig.~\ref{fig:orchestra_offload}.

\begin{table}[!t]
\centering
\caption{Breakdown of per-GPU activation offload volume per training step.}
\label{tab:offload-volume}
\footnotesize
\renewcommand{\arraystretch}{1.05}
\setlength{\tabcolsep}{4.5pt}
\begin{tabular}{ccrr}
\hline
\textbf{Patch/DMP/DP} & \textbf{$M_b$ (GB)} &
\multicolumn{2}{c}{\textbf{Offload Volume (GB)}} \\
\cline{3-4}
 & & \textbf{Sampling} & \textbf{Transformer} \\
\hline
2/4/2 & 40 &  9.7 & 16.7 \\
      & 30 & 72.5 & 29.9 \\
\hline
2/8/1 & 20 &  4.3 &  9.9 \\
      & 16 & 36.2 & 14.9 \\
\hline
4/1/8 & 80 & 22.8 & 10.8 \\
      & 65 & 22.8 & 27.5 \\
\hline
4/2/4 & 40 & 10.5 &  5.4 \\
      & 30 & 42.8 & 14.9 \\
\hline
\end{tabular}
\end{table}

\subsection{Forecasting Accuracy}\label{sec:exp_rmse}

\begin{figure*}[!t]
\centering
\subfloat[Pretraining with different patch and sampling configurations.]{
  \includegraphics[width=0.47\textwidth]{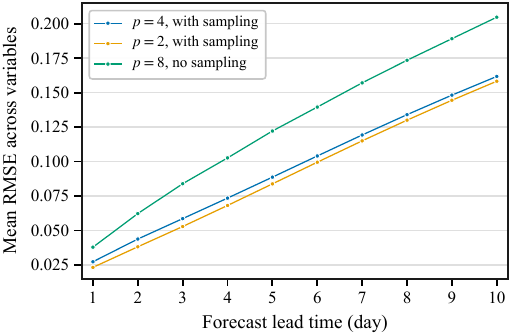}
  \label{fig:rmse_no_ft}
}
\hfill
\subfloat[Rollout finetuning with different rollout lengths.]{
  \includegraphics[width=0.47\textwidth]{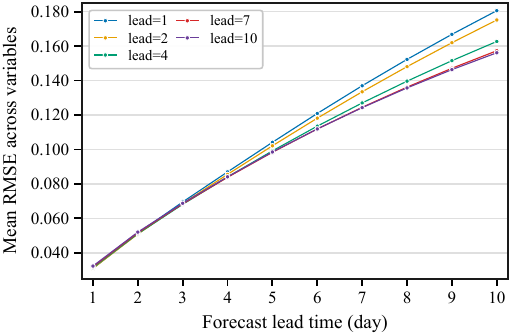}
  \label{fig:rmse_with_ft}
}
\caption{Mean RMSE of Wenhai forecasts at different lead times.}
\label{fig:exp_rmse}
\end{figure*}

\begin{figure*}[!t]
\centering
\includegraphics[width=\textwidth]{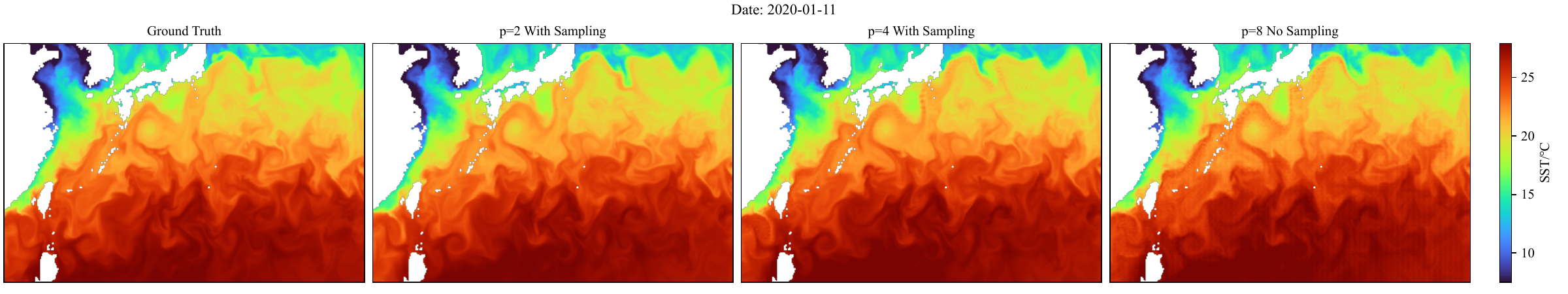}
\caption{Day-10 SST forecasts over the western North Pacific.}
\label{fig:sst_four_panel}
\end{figure*}

We finally evaluate the forecasting accuracy gains enabled by TERRA through hierarchical sampling, smaller patch sizes, and longer rollout finetuning. We train, finetune, and evaluate the small Wenhai model in Table~\ref{tab:scaling_models}. We use GLORYS data from 1993 to 2018 for pretraining and finetuning, and data from 2020 for validation. We pretrain each model for 200 epochs and finetune each rollout variant for 5000 steps. We report the mean RMSE over all ocean variables at each forecast lead time, where lower values indicate more accurate forecasts.

Fig.~\ref{fig:exp_rmse}\subref{fig:rmse_no_ft} compares pretrained models with different patch and sampling configurations. To match the number of tokens processed by the Transformer blocks, we use $p=8$ for the model without hierarchical sampling and $p=4$ with a down-sampling stride of $\alpha=2$ for the hierarchical model. The $p=2$ model with hierarchical sampling achieves the lowest RMSE at every evaluated forecast lead. With the sampling configuration fixed, reducing the patch size from $p=4$ to $p=2$ decreases the day-10 mean RMSE by $2.1\%$. Together, these results demonstrate the forecasting benefits of finer patch granularity and hierarchical sampling, highlighting the importance of TERRA's support for the parallel execution of sampling modules.

Fig.~\ref{fig:exp_rmse}\subref{fig:rmse_with_ft} fixes the $p=4$ model with hierarchical sampling and compares the pretrained model with variants finetuned using different rollout lengths. Increasing the rollout length consistently improves long-lead accuracy. In particular, ten-lead finetuning reduces the day-10 mean RMSE from $0.181$ for the pretrained model to $0.156$, corresponding to a $13.5\%$ reduction. The improvement becomes smaller beyond seven leads. Overall, these results demonstrate that the longer rollout finetuning enabled by TERRA improves long-range forecasting accuracy.

Fig.~\ref{fig:sst_four_panel} presents ten-day sea surface temperature (SST) forecasts initialized from the GLORYS global ocean state on January 1, 2020. It compares the forecasts valid on January 11, 2020, from the pretrained configurations in Fig.~\ref{fig:exp_rmse}\subref{fig:rmse_no_ft} with the corresponding GLORYS ground truth. The model without hierarchical sampling produces the most spatially smoothed forecast, with a substantial loss of fine-scale SST structures. With hierarchical sampling, the $p=2$ model resolves finer and clearer spatial details than the $p=4$ model, which is consistent with the lower day-10 mean RMSE achieved by the $p=2$ model.

\section{Conclusion}
We present TERRA in this paper, a hierarchical parallel training and memory orchestration framework for high-resolution AI-based Earth forecasting models. TERRA's SAWSTP parallelizes convolutional sampling modules and Swin Transformer blocks and connects them through differentiable token routes. Cost-aware window assignment reduces padding and shifted-window communication, while MO reduces the memory footprint of long-lead rollout finetuning. Experiments on Wenhai show that TERRA scales the 11.4B model to 96 H200 GPUs with $65.0\%$ strong-scaling efficiency. Compared with checkpoint-only policies, MO reduces peak allocated GPU memory by $32.2\%$--$51.8\%$ with at most $20.0\%$ step-time overhead. These capabilities make smaller patch sizes and longer rollout finetuning feasible, thereby improving long-lead forecasting accuracy.

\ifCLASSOPTIONcaptionsoff
  \newpage
\fi

\bibliographystyle{IEEEtran}
\bibliography{refer}

\end{document}